\PassOptionsToPackage{frozencache}{minted}

\IfFileExists{./prepreamble-amspreprint.sty}{\RequirePackage[packages,theorems,changes]{prepreamble-amspreprint}}{}

\makeatletter
\IfFileExists{./scoop-latex/scoop-packages.sty}{\providecommand*{\input@path}{}\edef\input@path{{./scoop-latex/}\input@path}}{}
\makeatother
\PassOptionsToPackage{style=authoryear-comp,backend=biber}{biblatex}
\PassOptionsToPackage{commentmarkup = todo}{changes}    

\documentclass[english]{amsart}

\RequirePackage{biblatex}
\RequirePackage{ltxcmds}
\IfFileExists{./preamble-amspreprint.sty}{\RequirePackage[packages,theorems,changes]{preamble-amspreprint}}{}

\usepackage{scoop-local}

\IfFileExists{./postpreamble-amspreprint.sty}{\RequirePackage[packages,theorems,changes]{postpreamble-amspreprint}}{}

\makeatletter
\@ifpackageloaded{changes}{
\definechangesauthor[name = {Leonie Kreis}, color = {red!80!black}]{LK}
\definechangesauthor[name = {Evelyn Herberg}, color = {blue!80!black}]{EH}
\definechangesauthor[name = {Frederik Köhne}, color = {orange!80!black}]{FK}
\definechangesauthor[name = {Anton Schiela}, color = {green!70!black}]{AS}
\definechangesauthor[name = {Roland Herzog}, color = {teal}]{RH}
}{}
\makeatother        

\makeatletter
\@ifpackageloaded{hyperref}{%
	\hypersetup{
		pdftitle = {SensLI: Sensitivity-Based Layer Insertion for Neural Networks},
		pdfauthor = {Leonie Kreis, Evelyn Herberg, Frederik Köhne, Anton Schiela, Roland Herzog},
		pdfkeywords = {deep learning, adaptive network architecture, layer insertion, sensitivity analysis},
		pdfcreator = {Created using the Scoop Template Engine version 1.6.0.}
	}
}{
	\pdfinfo{
		/Title (SensLI: Sensitivity-Based Layer Insertion for Neural Networks)
		/Author (Leonie Kreis, Evelyn Herberg, Frederik Köhne, Anton Schiela, Roland Herzog)
		/Subject ()
		/Keywords (deep learning, adaptive network architecture, layer insertion, sensitivity analysis)
		/Creator (Created using the Scoop Template Engine version 1.6.0.)
	}
}
\makeatother

\title[SensLI: Sensitivity-Based Layer Insertion]{SensLI: Sensitivity-Based Layer Insertion for Neural Networks}

\author[L. Kreis]{Leonie Kreis\orcidlink{0000-0003-4234-4867}}
\address[L. Kreis]{Interdisciplinary Center for Scientific Computing, Heidelberg University, 69120 Heidelberg, Germany}
\email{\detokenize{leonie.kreis@iwr.uni-heidelberg.de}}
\urladdr{https://scoop.iwr.uni-heidelberg.de}

\author[E. Herberg]{Evelyn Herberg\orcidlink{0000-0003-2515-4818}}
\address[E. Herberg]{Interdisciplinary Center for Scientific Computing, Heidelberg University, 69120 Heidelberg, Germany}
\email{\detokenize{evelyn.herberg@iwr.uni-heidelberg.de}}
\urladdr{https://scoop.iwr.uni-heidelberg.de}

\author[F. Köhne]{Frederik Köhne\orcidlink{0009-0008-6185-9675}}
\address[F. Köhne]{Department of Mathematics, University of Bayreuth, 95440 Bayreuth, Germany}
\email{\detokenize{frederik.koehne@uni-bayreuth.de}}
\urladdr{https://num.math.uni-bayreuth.de/en/team/frederik-koehne/}

\author[A. Schiela]{Anton Schiela\orcidlink{0000-0002-6959-2951}}
\address[A. Schiela]{Department of Mathematics, University of Bayreuth, 95440 Bayreuth, Germany}
\email{\detokenize{anton.schiela@uni-bayreuth.de}}
\urladdr{https://num.math.uni-bayreuth.de/en/team/anton-schiela/}

\author[R. Herzog]{Roland Herzog\orcidlink{0000-0003-2164-6575}}
\address[R. Herzog]{Interdisciplinary Center for Scientific Computing, Heidelberg University, 69120 Heidelberg, Germany}
\email{\detokenize{roland.herzog@iwr.uni-heidelberg.de}}
\urladdr{https://scoop.iwr.uni-heidelberg.de}

\date{\today}

\dedicatory{}

\begin{document}

\begin{abstract}
The training of neural networks requires tedious and often manual tuning of the network architecture.
We propose a systematic approach to inserting new layers during the training process.
Our method eliminates the need to choose a fixed network size before training, is numerically inexpensive to execute and applicable to various architectures including fully connected feedforward networks, ResNets and CNNs.
Our technique borrows ideas from constrained optimization and is based on first-order sensitivity information of the loss function with respect to the virtual parameters that additional layers, if inserted, would offer.
In numerical experiments, our proposed sensitivity-based layer insertion technique (SensLI) exhibits improved performance on training loss and test error, compared to training on a fixed architecture, and reduced computational effort in comparison to training the extended architecture from the beginning.
Our code is available on \url{https://github.com/mathemml/SensLI}.

\end{abstract}


\makeatletter
\ltx@ifpackageloaded{hyperref}{%
\subjclass[2010]{}
}{%
\subjclass[2010]{}
}
\makeatother

\maketitle

\section{Introduction}
\label{section:introduction}

Suitable network architectures for supervised learning are, in general, a-priori unknown.
Hence in practice, the architecture is often chosen by experience and experiment.
The field of neural architecture search (NAS) has arisen in order to tackle this issue by finding a suitable architecture prior to the actual training process.
However, NAS algorithms are generally considered computationally expensive.

\subsection{Our Contribution}

We present a general framework for sensitivity-based layer insertion (SensLI) during training.
SensLI is a depth-adaptive method that can be viewed as automated hyperparameter search for the depth of the network.
An exemplary schematic visualization is shown in \cref{figure:FNN-CNN-positions-of-new-layers}.

Specifically, SensLI is based on the sensitivity of the loss function's current value with respect to the virtual weights associated with all potential new layers.
It requires only moderate computational effort and is applicable to various architectures including fully connected feedforward neural networks (FNNs), residual neural networks (ResNets) and convolutional neural networks (CNNs).

\begin{figure}[t]
	\centering
	\includegraphics[width = \linewidth]{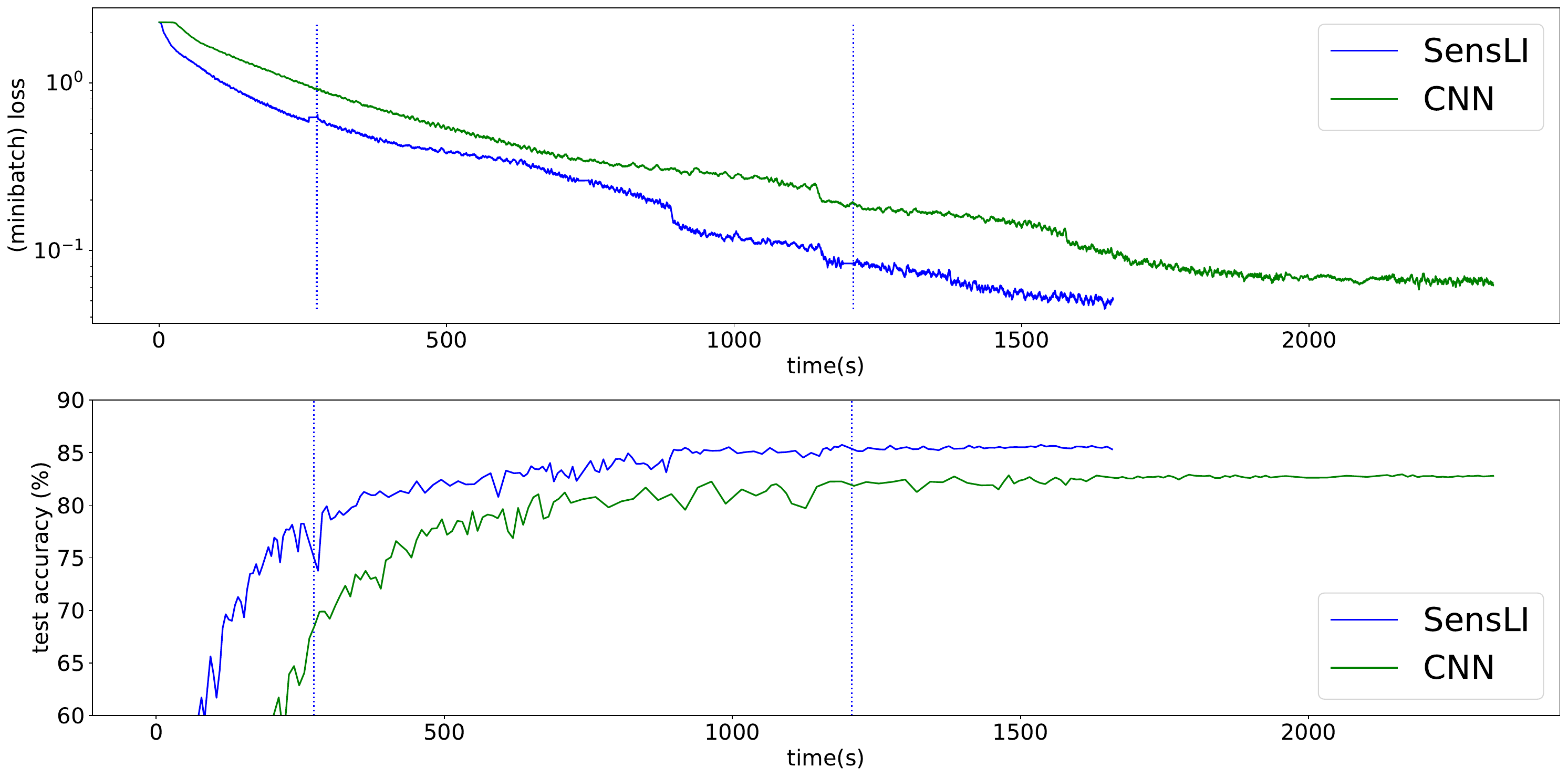}
	\caption{%
		Training a CNN on the CIFAR-10 data set with multiple layer insertions (SensLI), compared to training the extended CNN from the beginning, plotted over time in seconds.
		We display the training loss (top) and test accuracy (bottom).
		SensLI is executed every 50~epochs, \ie, 3~times throughout the training run, but decides against the second layer insertion, because the threshold in \eqref{eq:when-to-insert} was not met. Hence, only 2 layer insertions take place, which are indicated by vertical lines.
		This experiment is included in the \github repository as Exp1 and the detailed experiment setup is documented in \cref{section:hyperparameters}.
	}
	\label{figure:performance-of-layer-insertion-cnn}
\end{figure}

\Cref{figure:performance-of-layer-insertion-cnn} shows that growing a CNN architecture during training using SensLI can outperform training the extended architecture from the beginning.
Additionally, due to smaller networks during training, the computational effort is reduced, which also results in a shorter training time.

We describe SensLI in \cref{section:automated-layer-insertion}, provide numerical results in \cref{section:numerical-results} and close with concluding remarks in \cref{section:discussion}.

\subsection{Related Work}

Modifying the architecture of a neural network during training is not a new idea.
It dates back to methods such as cascade correlation proposed in \citet{FahlmannLebiere:1989:1} and the RCE network proposed in \citet{Hudak:1991:1}.
There are two main ways to expand a given network: by inserting layers, or by widening existing layers.

Methods in the literature differ with respect to how, where and when they insert new neurons into the architecture.
Additionally, they vary with regard to their computational cost and which architectures they can be applied to.
Layer widening is more thoroughly explored in the literature compared to layer insertion.
We provide a tabular overview of relevant literature in \cref{tab:comparison}.

Net2Net proposed in \citet{ChenGoodfellowShlens:2015:1} and NetMorph proposed in \citet{WeiWangRuiChen:2016:1} handle the question of how to initialize new neurons effectively after layer insertion.
Gradmax proposed in \citet{EvciVanMerrienboerUnterthinerVladymyrovPedregosa:2022:1} concentrates on initialization strategies for layer widening using singular value decomposition, prioritizing training dynamics over immediate decrease in the objective function.
AutoGrow proposed in \citet{WenYanChenLi:2020:1} focuses on automatization instead of acceleration of training and compares different empirical initialization and trigger strategies.
The Firefly architecture descent from \citet{WuLiuStoneLiu:2020:1} and splitting steepest descent from \citet{WuWangLiu:2019:1} propose methods that jointly optimize network parameters and architecture in an alternating fashion.
NeST proposed in \citet{DaiYinJha:2019:1} suggests a method for layer widening with gradient-based growth of neurons and magnitude-based pruning.
MorphNet proposed in \citet{GordonEbanNachumChenWuYangChoi:2018:1} shrinks and grows the network during training, subject to resource constraints.
Finally, the strategies proposed in \citet{MaileRachelsonLugaWilson:2022:1} make use of orthogonal weight initialization and triggers.

While the above methods are mainly heuristic in nature, some approaches rest on more analytical foundations.
\citet{VerbockhavenChevallierCharpiat:2023:1} and \citet{MitchellMenzenbachKerstingMundt:2023:1} use information of the objective function and the natural gradient respectively.
AdaNet proposed in \citet{CortesGonzalvoKuznetsovMohriYang:2017:1} learns network structures and weights simultaneously based on data-driven theoretical guarantees.
For residual neural networks, automated layer insertion leveraging the neural ODE perspective is considered, \eg, \citet{ChangMengHaberTungBegert:2018:1} and \citet{DongLiuLiShang:2020:1}.

\begin{table}[H]
	\caption{%
		Comparison of methods for expanding neural networks.
		Indicating whether layer insertion or widening was considered, whether the question on how, where and when to insert were answered (\cf \cref{subsection:layer-insertion-during-training}), whether the network expansion was executed during training and which architectures were examined out of FNN, ResNet and CNN.
	}
	\label{tab:comparison}
	\resizebox{\textwidth}{!}{
		\begin{tabular}{lcccccccccc}
			\toprule
			& \multicolumn{2}{c}{\textbf{Layer}} & & & & \textbf{During} & \multicolumn{3}{c}{\textbf{Architecture}}
			\\
			\cmidrule{2-3}
			\cmidrule{8-10}
			\textbf{Method} & \textbf{Insertion} & \textbf{Widening} & \textbf{How?} & \textbf{Where?} & \textbf{When?} & \textbf{training?}& \textbf{FNN} & \textbf{ResNet} & \textbf{CNN}
			\\
			\midrule
			Net2Net \citep{ChenGoodfellowShlens:2015:1}                              & \yes & \yes & \yes & \no  & \no  & \no  & \yes & \no  & \yes &
			\\
			\midrule
			NetMorph \citep{WeiWangRuiChen:2016:1}                                   & \yes & \yes & \yes & \no  & \no  & \no  & \yes & \no  & \yes &
			\\
			\midrule
			Firefly \citep{WuLiuStoneLiu:2020:1}                                     & \yes & \yes & \yes & \yes & \no  & \yes & \yes & \yes  & \yes &
			\\
			\midrule
			Autogrow \citep{WenYanChenLi:2020:1}                                     & \yes & \no  & \yes & \yes & \yes & \yes & \no  & \no  & \yes &
			\\
			\midrule
			SENN \citep{MitchellMenzenbachKerstingMundt:2023:1}                      & \yes & \yes & \yes & \yes & \yes & \yes & \yes & \yes & \no  &
			\\
			\midrule
			ConvSENN \citep{DeaconuAppolinaryYangLi:2024:1}                          & \yes & \yes & \yes & \yes & \yes & \yes & \no  & \no  & \yes &
			\\
			\midrule
			MorphNet \citep{GordonEbanNachumChenWuYangChoi:2018:1}                   & \no  & \yes & \no  & \no  & \yes & \no  & \yes & \yes & \yes &
			\\
			\midrule
			NeST \citep{DaiYinJha:2019:1}                                            & \no  & \yes & \yes & \yes & \yes & \yes & \yes & \yes & \yes &
			\\
			\midrule
			Splitting \citep{WuWangLiu:2019:1}                                       & \no  & \yes & \yes & \yes & \yes & \no  & \yes & \no  & \no  &
			\\
			\midrule
			GradMax \citep{EvciVanMerrienboerUnterthinerVladymyrovPedregosa:2022:1}  & \no  & \yes & \yes & \no  & \no  & \yes & \yes & \no  & \yes &
			\\
			\midrule
			\citep{MaileRachelsonLugaWilson:2022:1}                                  & \no  & \yes & \yes & \yes & \yes & \yes & \yes & \no  & \no  &
			\\
			\midrule
			\citep{VerbockhavenChevallierCharpiat:2023:1}                            & \no & \yes  & \yes & \yes & \yes & \yes & \yes & \yes & \yes &
			\\
			\midrule
			\textbf{SensLI} (this paper)                                            & \yes & \no  & \yes & \yes & \yes & \yes & \yes & \yes & \yes &
			\\
			\bottomrule
		\end{tabular}
	}
\end{table}

\section{SensLI: Sensitivity-Based Layer Insertion}
\label{section:automated-layer-insertion}

In this section we describe the proposed SensLI strategy and begin by establishing notation.

\subsection{Notation for Feedforward and Residual Neural Networks}
\label{subsection:notation}

We consider supervised deep learning problems of the form
\begin{equation}
	\label{eq:classical-training}
	\text{Minimize}
	\quad
	\frac{1}{N} \sum_{i=1}^N \ell(\network{H}(\theta)(x_i),y_i)
\end{equation}
with trainable parameters~$\theta$ from some vector space.
The pairs $(x_i,y_i)_{i = 1, \ldots, N}$ are the training data with features $x_i \in \R^d$ and labels $y_i \in \R^c$.
The function $\ell \colon \R^c \times \R^c \to \R$ denotes the loss function, and $\network{H}(\theta) \colon \R^d \to \R^c$ is the propagation function represented by the neural network under consideration, given the parameters~$\theta$.

In case of a feedforward neural network (FNN) with $L$~hidden layers and activation function $\sigma \colon \R \to \R$ applied componentwise, $\network{H}(\theta)$ has the decomposed form
\begin{subequations}
	\label{eq:FNN}
	\begin{align}
		x^k
		&
		=
		\sigma(W^k x^{k-1} + b^k)
		\quad
		\text{for }
		k = 1, \ldots, L
		,
		\label{eq:FNN:1}
		\\
		x^{L+1}
		&
		=
		W^{L+1} x^L + b^L
		.
		\label{eq:FNN:2}
	\end{align}
\end{subequations}
The vector $x^0 \in \R^d = \R^{h_0}$ denotes the network's input, $x^{L+1} \in \R^c = \R^{h_{L+1}}$ is its output, and the remaining $x^k \in \R^{h_k}$ are quantities on intermediate layers.
The trainable parameters $\theta$ comprise the weight matrices $W^k \in \R^{h_k \times h_{k-1}}$ and bias vectors $b^k \in \R^{h_k}$.

In case of a residual neural network (ResNet), we consider an architecture inspired by \citet{HaberRuthotto:2017:2} with propagation function $\network{H}(\theta)$ of the decomposed form
\begin{subequations}
	\label{eq:ResNet}
	\begin{align}
		x^1
		&
		=
		W^1 x^0
		\label{eq:ResNet:1}
		,
		\\
		x^k
		&
		=
		x^{k-1}
		+
		W_2^k \, \sigma(W_1^k x^{k-1} + b^k)
		,
		\label{eq:ResNet:2}
		\\
		x^{L+1}
		&
		=
		W^{L+1} x^L
		\label{eq:ResNet:3}
	\end{align}
\end{subequations}
for all $k = 2, \ldots, L$.
All intermediate weight matrices $W_1^k, W_2^k \in \R^{h_k \times h_k}$ are square for $k = 2, \ldots, L$ with widths $h_1 = \ldots = h_L$, while the initial and terminal weights have dimensions $W^1 \in \R^{h_1 \times h_0}$ and $W^{L+1} \in \R^{h_{L+1} \times h_L}$.
The bias vectors $b^k \in \R^{h_k}$ are also all of the same dimension.

\subsection{Notation for Convolutional Neural Networks}
\label{subsection:notation-cnn}

\begin{figure}[t]
	\centering
	\includegraphics[width = \linewidth]{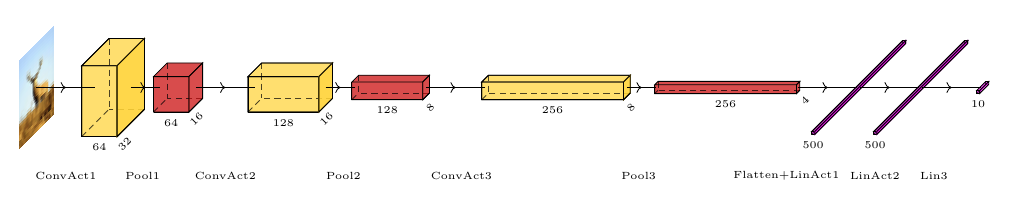}
	\caption{Illustration of a VGG-inspired CNN architecture.}
	\label{figure:CNN-architecture}
\end{figure}

We consider VGG-inspired CNN architectures \citep{SimonyanZisserman:2014:2} with $L_1$ convolutional layers and $L_2$ fully connected layers.
The baseline CNN architecture we use is shown in \cref{figure:CNN-architecture}.

The propagation function $\network{H}(\theta)$ of the CNN has the decomposed form
\begin{subequations}
	\label{eq:CNN}
	\begin{align}
		&
		Y^k
		=
		\sigma(K^k \convolution X^{k-1} + b^k)
		\quad
		\text{for }
		k = 1, \ldots, L_1
		,
		\label{eq:CNN:1}
		\\
		&
		X^k
		=
		\text{MaxPool}(Y^k)
		\quad
		\text{for }
		k = 1, \ldots, L_1
		,
		\label{eq:CNN:2}
		\\
		&
		x^{L_1}
		=
		\textup{Flatten}(X^{L_1})
		,
		\\
		&
		x^{k+1}
		=
		\sigma(W^k x^k + b_{\text{fc}}^k)
		\quad
		\text{for }
		k = L_1, \ldots, L_1 + L_2 - 2
		,
		\label{eq:CNN:3}
		\\
		&
		x^{L_1+L_2}
		=
		W^{L_1+L_2} x^{L_1+L_2-1}
		\label{eq:CNN:4}
		.
	\end{align}
\end{subequations}
The tensor $X^0 \in \R^{m_0 \times d_0 \times d_0} $ denotes the network's input with $m_0$ being the number of channels and $d_0 \times d_0$ the spatial dimensions of each channel.
$Y^k \in \R^{m_k \times d_{k-1} \times d_{k-1}}$ are intermediate layers after the activation and $X^k\in\R^{m_k \times d_k \times d_k}$ with $d_k \coloneqq \frac{d_{k-1}}{2}$.
Here, $\convolution$ denotes convolution, and the convolutional kernels $K^k$ use padding and stride values of~1 so that the spatial dimensions of $X^{k-1}$ are preserved.
The bias vectors in the convolutional layers are applied channel-wise and the activation function is applied element-wise.
We use ReLU as activation function.
The operation $\text{MaxPool}$ is the max-pooling operation with a kernel size of $2 \times 2$ and stride~$2$.
The operation $\textup{Flatten}$ transforms the tensor of dimension $m_{L_1} \times d_{L_1 -1} \times d_{L_1-1}$ to a vector of dimension $m_{L_1} d_{L_1-1}^2=:d_{L_1}$; then $x^k \in\R^{d_k}$ for all $k = L_1, \ldots, L_1 + L_2 - 1$ are intermediate values of the fully connected layers and $x^{L_1+L_2} \in \R^{d_{L_1+L_2}} = \R^c $ is its output.
The trainable parameters $\theta$ comprise the kernels $K^k \in \R^{3 \times 3 \times m_k \times m_{k-1}}$, the bias vectors $b^k \in \R^{m_k}$ and the weight matrices $W^k \in \R^{d_{k+1} \times d_k}$ and bias vectors $b_{\text{fc}}^k \in \R^{d_{k+1}}$ of the fully connected layers.

With either architecture, the training problem, \eqref{eq:classical-training}, is an unconstrained, usually nonconvex optimization problem, where the smoothness of the objective depends on the smoothness of the activation function~$\sigma$ and of the loss function~$\ell$.
Popular choices of optimizers to train a network are mini-batch stochastic gradient descent (mini-batch SGD); see, \eg, \citet{BilmesAsanovicChinDemmel:1997:1,Bottou:2010:1,BottouCurtisNocedal:2018:1}, and the Adam optimizer \citep{KingmaBa:2015:1}.

\subsection{Layer Insertion During Training}
\label{subsection:layer-insertion-during-training}

In classical training approaches to solving the training problem, \eqref{eq:classical-training}, the network architecture and dimensions are selected prior to the training process and are kept fixed during the training.
In the following we look at a technique to insert a new layer into a network architecture during the training process.
Three main questions arise:
\begin{itemize}
	\item
		\textbf{Where} to insert the new layer in the network?

	\item
		\textbf{How} to initialize the parameters of the new layer?

	\item
		\textbf{When} to insert the new layer during training?
\end{itemize}
In this contribution, we focus on answering the first and second question, borrowing ideas from sensitivity analysis in constrained optimization for the former.
The proposed technique is compatible with any optimizer used for network training.
Additionally, we propose a heuristic to answer the third question.

\subsection{Initialization of a New Layer}
\label{subsection:initialization}

Suppose that a new hidden layer is inserted between two existing layers in a feedforward or residual baseline neural network at some point during training.
This results in additional trainable parameters~$\thetanew$ being added to the network, which then become part of the extended set of parameters $\thetaextended = [\thetaold, \thetanew]$.

Clearly, the goal of layer insertion is to allow the extended network to represent a richer space of functions than the baseline network.
In order to take full advantage of this, we initialize the newly added trainable parameters $\thetanew$ with two goals in mind:
\begin{enumeratearabic}
	\item \label[property]{item:extension-goals:1}
		The propagation function $\networkextended(\thetaextended)$ of the extended network should initially be identical to the propagation function $\networkold(\thetaold)$ of the baseline network at the current point $\thetaold$ in training.

	\item \label[property]{item:extension-goals:2}
		The loss function $\ell(\networkextended(\thetaextended)(x_i),y_i)$ at a typical data point $(x_i,y_i)$ should have non-zero gradient components \wrt $\thetanew$ after initialization.
		As a consequence, the training algorithm can make use of the new parameters immediately.
\end{enumeratearabic}
For ResNets as in \eqref{eq:ResNet}, it is straightforward to ensure \cref{item:extension-goals:1} by initializing the propagation realized by the newly added layer to be the identity function.
Suppose we insert a layer after the $k$-th layer and initialize its parameters $\thetanew = [W_1, W_2, b]$ using
\begin{equation}
	\label{eq:ResNet:initialization}
	\begin{aligned}
		&
		W_1
		\in
		\R^{h_k \times h_k}
		\text{ and }
		b
		\in
		\R^{h_k}
		\text{ arbitrary}
		,
		\\
		&
		W_2
		\coloneqq
		0
		\in
		\R^{h_k \times h_k}
		.
	\end{aligned}
\end{equation}

Then the partial propagation function~$\networkaddedlayer$ realized by the inserted layer is indeed the identity:
\begin{align*}
	x^+
	&
	\coloneqq
	\networkaddedlayer(\thetanew)(x^{k-1})
	=
	x^{k-1}
	+
	W_2 \, \sigma(W_1 x^{k-1} + b)
	\\
	&
	=
	x^{k-1}
	+
	0 \, \sigma(W_1 x^{k-1} + b)
	=
	x^{k-1}
	.
\end{align*}

Abbreviating $\fextended^{(i)}(\thetaextended) \coloneqq \ell(\networkextended(\thetaextended)(x_i),y_i)$ for a data point $(x_i,y_i)$, a straightforward calculation shows
\begin{subequations}
	\label{eq:ResNet:initialization:gradients}
	\begin{align}
		\nabla_{W_1} \fextended^{(i)}(\thetaextended)
		&
		=
		0
		,
		\label{eq:ResNet:initialization:gradients:1}
		\\
		\nabla_b \fextended^{(i)}(\thetaextended)
		&
		=
		0
		,
		\label{eq:ResNet:initialization:gradients:2}
		\\
		\nabla_{W_2} \fextended^{(i)}(\thetaextended)
		&
		=
		\nabla_{x_i^k} \fextended^{(i)}(\thetaextended) \, \sigma(W_1 x_i^{k-1} + b)^\transp
		,
		\label{eq:ResNet:initialization:gradients:3}
	\end{align}
\end{subequations}
where gradients, as usual, are evaluated with respect to the Euclidean and Frobenius inner products, respectively.
Moreover, $x_i^k$ is the value of the input~$x_i$, propagated to the $k$-th hidden layer.
For a fair comparison of all potential layer positions it is recommended to initialize $W_1$ in the same range (or even as the same matrix) for all layers, so that it does not impact the magnitudes of $\nabla_{W_2} \fextended^{(i)}(\thetaextended)$, \cf \eqref{eq:ResNet:initialization:gradients:3}.

With \cref{item:extension-goals:2} of the loss having non-zero gradient \wrt the additional parameters in mind, a further requirement on the initialization in \eqref{eq:ResNet:initialization} arises.
From \eqref{eq:ResNet:initialization:gradients:3}, we infer that we need to initialize the new weight~$W_1$ and bias~$b$ such that $\sigma(W_1 x + b)$ is not systematically zero.
For instance, for $\sigma = \tanh$ or leaky \relu, we can choose $W_1 = \id$ and $b = 0$.
Note that the choice of $W_1$ will determine the scale or norm of the resulting gradient $\nabla_{W_2} \fextended(\thetaextended)$.
We also infer from \eqref{eq:ResNet:initialization:gradients} that the inner weight matrix~$W_1$ and bias vector~$b$ will only start to receive non-zero updates from the second (mini-batch) gradient step onwards.

For feedforward \relu-networks as in \eqref{eq:FNN}, we can use the initialization from \citet{ChenGoodfellowShlens:2015:1} to insert a layer anywhere after the first hidden layer ($k \ge 2$).
This amounts to the choice
\begin{equation}
	\label{eq:FNN:initialization}
	W
	\coloneqq
	\id_{h_k \times h_k}
	,
	\quad
	b
	\coloneqq
	0
	\in
	\R^{h_k}
\end{equation}
for the new parameters $\thetanew = [W, b]$.
Notice that we choose the newly added layer to have the same width as its predecessor.
The initialization in \eqref{eq:FNN:initialization} results in the identity function since the \relu activation satisfies $\sigma \circ \sigma = \sigma$ and thus
\begin{align*}
	x^+
	&
	\coloneqq
	\networkaddedlayer(\thetanew)(x^{k-1})
	=
	\relu(\id_{h_k \times h_k} x^{k-1} + 0)
	\\
	&
	=
	\relu(x^{k-1})
	=
	x^{k-1}
\end{align*}
holds.
The last equality is true because $x^{k-1}$ itself is the output of the previous \relu layer and thus a vector with non-negative components.
Looking at the gradients of the new layer and using the notation from \eqref{eq:ResNet:initialization:gradients}, we obtain
\begin{subequations}
	\label{eq:FNN:initialization:gradients}
	\begin{align}
		\nabla_W \fextended^{(i)}(\thetaextended)
		&
		=
		(W^{k+1})^\transp \, \nabla_{W^{k+1}} \fextended^{(i)}(\thetaextended)
		,
		\label{eq:FNN:initialization:gradients:1}
		\\
		\nabla_b \fextended^{(i)}(\thetaextended)
		&
		=
		(W^{k+1})^\transp \, \nabla_{b^{k+1}} \fextended^{(i)}(\thetaextended)
		.
		\label{eq:FNN:initialization:gradients:2}
	\end{align}
\end{subequations}
Thus \eqref{eq:FNN:initialization:gradients} implies that upon initialization, we can expect a non-zero gradient \wrt the additional parameters.
To realize an identity layer initialization in FNNs with general activation functions, different from \relu, the activation function may need to be parametrized \citep{WeiWangRuiChen:2016:1}.

In the case of CNNs as in \eqref{eq:CNN}, we also concentrate on the \relu activation function.
We need to find an initialization of the kernel $K\in \R^{3 \times 3 \times m \times m}$ and the bias $b\in\R^m$ so that
\begin{equation*}
	\label{eq:CNN:identitypropagation}
	K \convolution X + b
	=
	X
	.
\end{equation*}
For all $i, j = 1, 2, 3$ and $k, \ell = 1, \ldots, m$, we set
\begin{subequations}
	\label{eq:CNN:initialization}
	\begin{align}
		K_{i,j,k,\ell}
		&
		\coloneqq
		\begin{cases}
			1
			&
			\text{if } i = j = 2 \text{ and } k = \ell
			,
			\\
			0
			&
			\text{otherwise}
			,
		\end{cases}
		\\
		b_i
		&
		\coloneqq
		0
		\quad
		\text{for }
		i = 1, \ldots, m
		.
	\end{align}
\end{subequations}
Then the forward propagation through the layer is indeed the identity function, as can be seen by direct calculation as for the FNN, \cf \eqref{eq:FNN:initialization}.

\begin{figure}[htp]
	\centering
	\includegraphics[width = 0.3\linewidth]{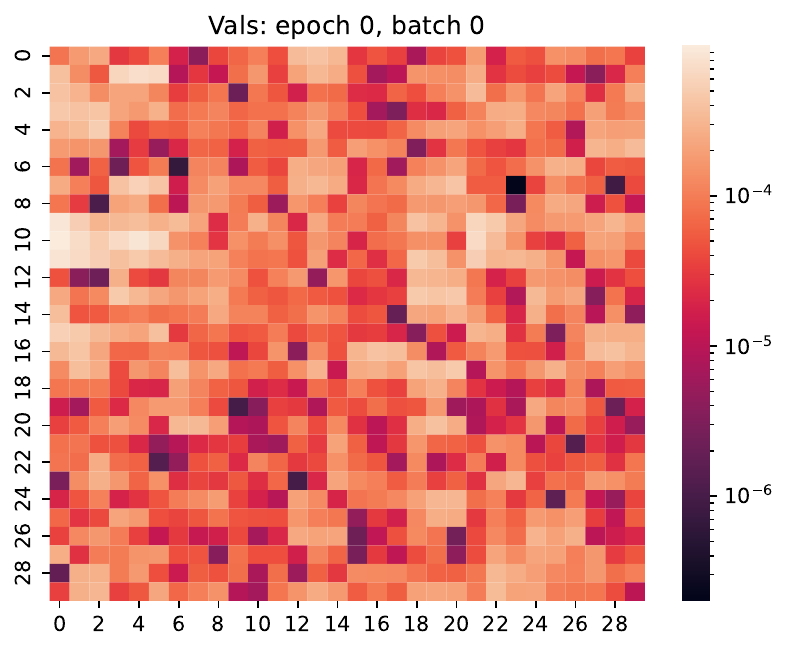}%
	\hfill
	\includegraphics[width = 0.3\linewidth]{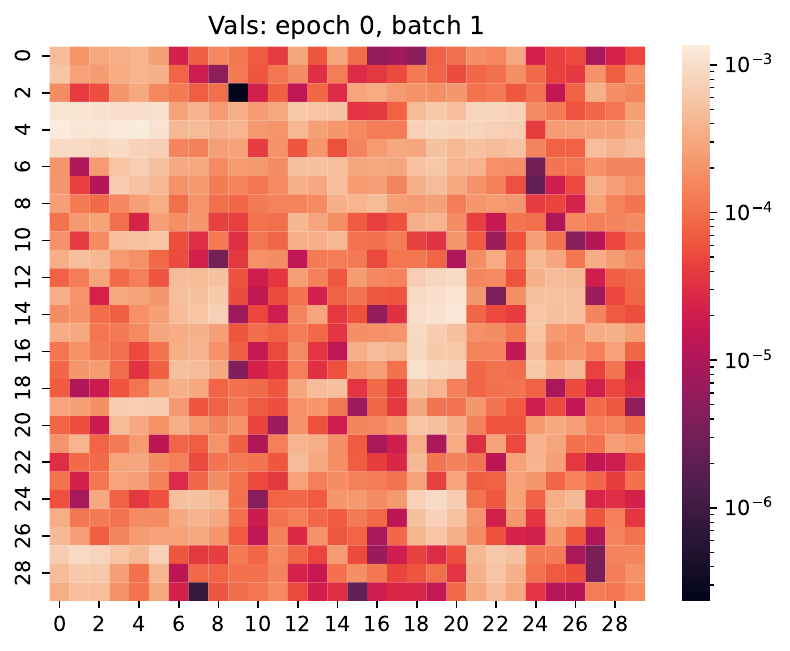}%
	\hfill
	\includegraphics[width = 0.3\linewidth]{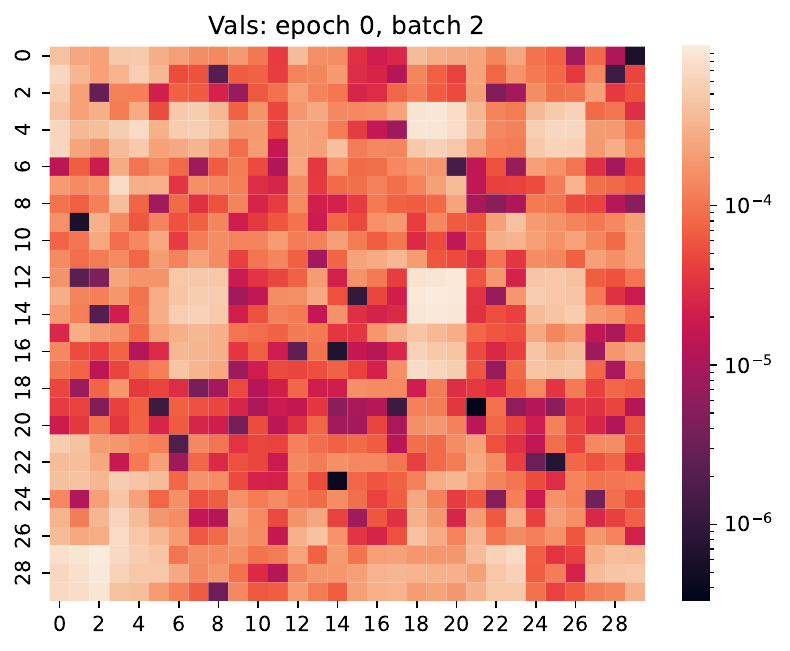}
	\caption{%
		Heatmaps showing the entries of the (Euclidean) gradient of a convolutional kernel, flattened to 2~dimensions, for the first three iterations after insertion.
		The kernel is newly inserted into the CNN following the initialization described in \eqref{eq:CNN:initialization}.
		Mini-batch SGD is used as optimizer.
	}
	\label{figure:heatmaps-of-grads-of-new-weight-cnn}
\end{figure}

\begin{figure}[htp]
	\centering
	\includegraphics[width = 0.3\linewidth]{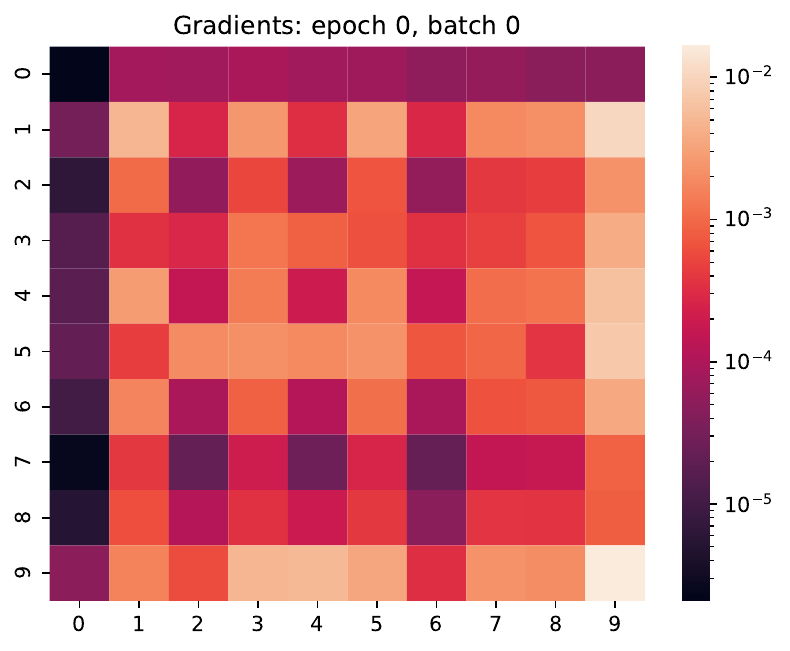}%
	\hfill
	\includegraphics[width = 0.3\linewidth]{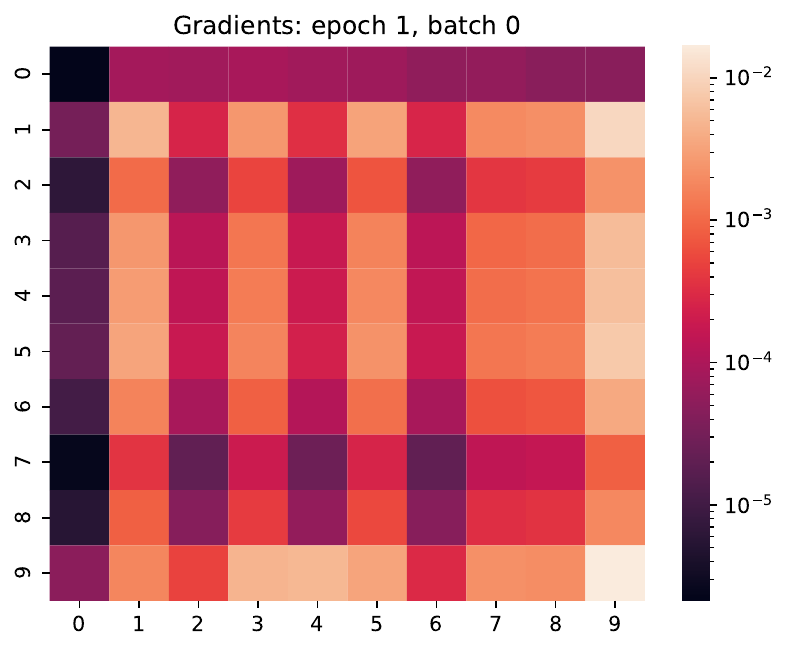}%
	\hfill
	\includegraphics[width = 0.3\linewidth]{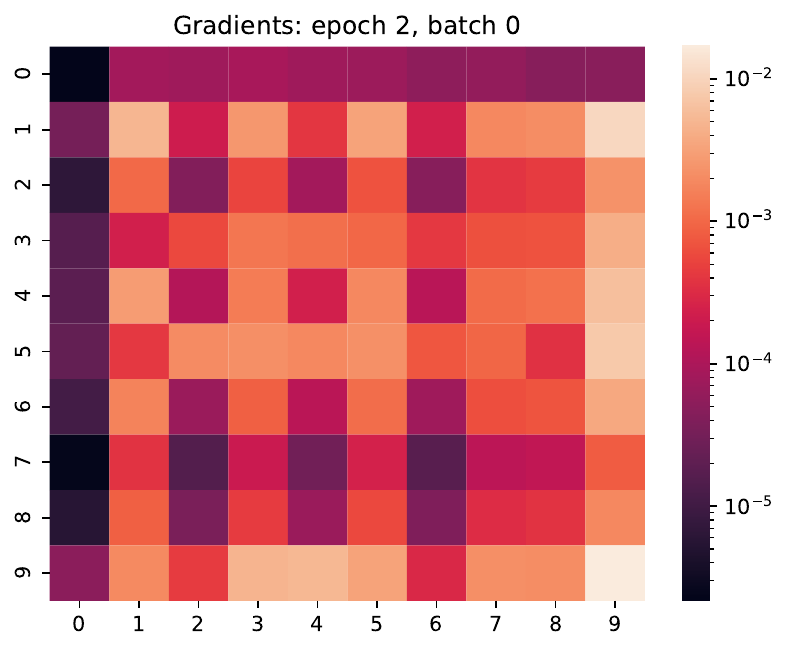}%
	\\
	\includegraphics[width = 0.3\linewidth]{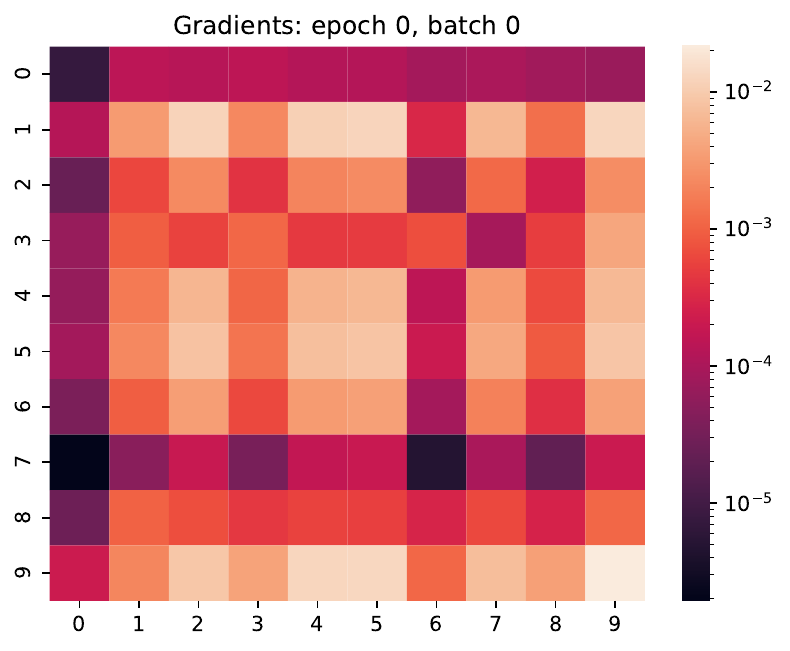}%
	\hfill
	\includegraphics[width = 0.3\linewidth]{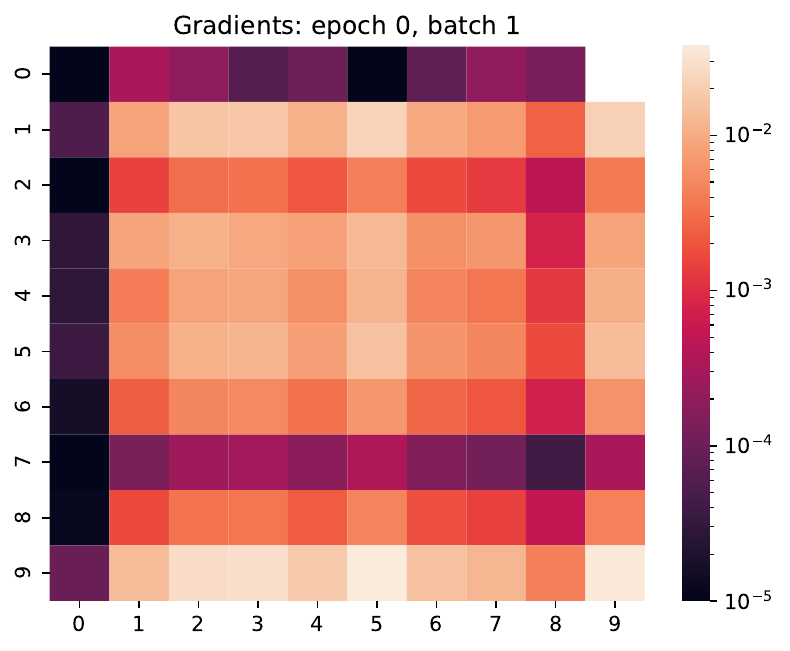}%
	\hfill
	\includegraphics[width = 0.3\linewidth]{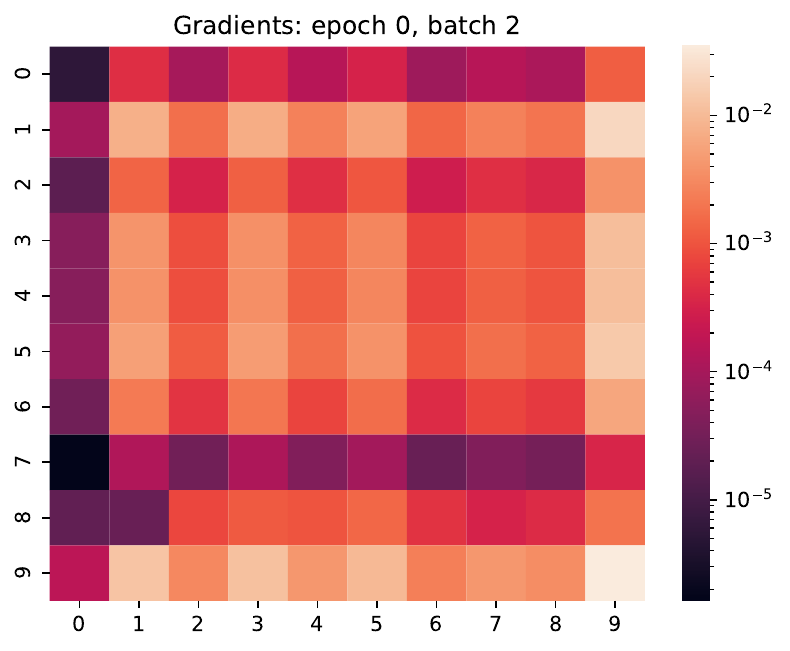}
	\caption{%
		Heatmaps showing the entries of the (Euclidean) gradient of a weight matrix, for the first three iterations after insertion.
		The weight matrix is newly inserted into a \relu-FNN following the initialization described in \eqref{eq:FNN:initialization}.
		Top row: full-batch training.
		Bottom row: mini-batch training.
	}
	\label{figure:heatmaps-of-grads-of-new-weight}
\end{figure}

\begin{figure}[htp]
	\centering
	\includegraphics[width = 0.3\linewidth]{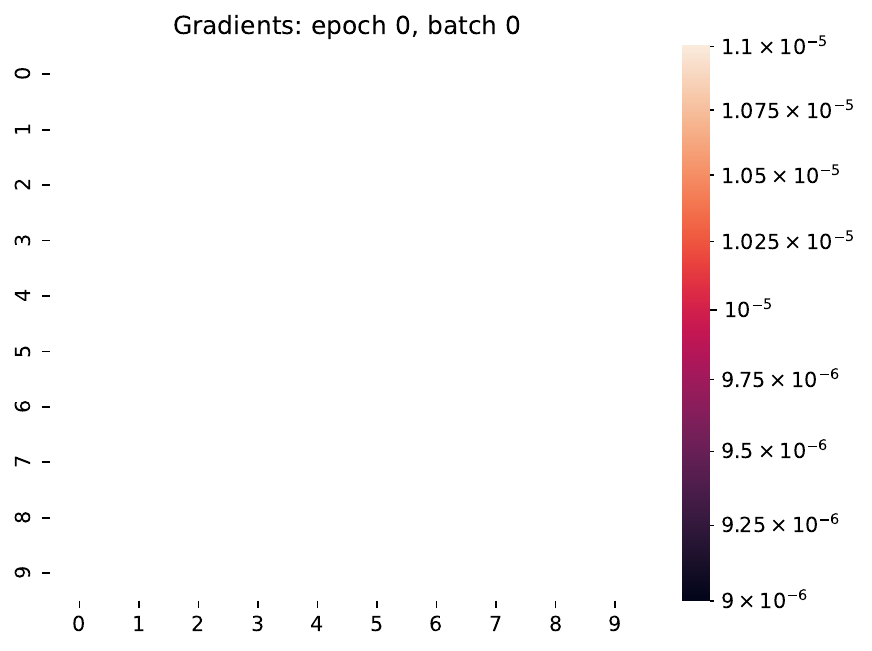}%
	\hfill
	\includegraphics[width = 0.3\linewidth]{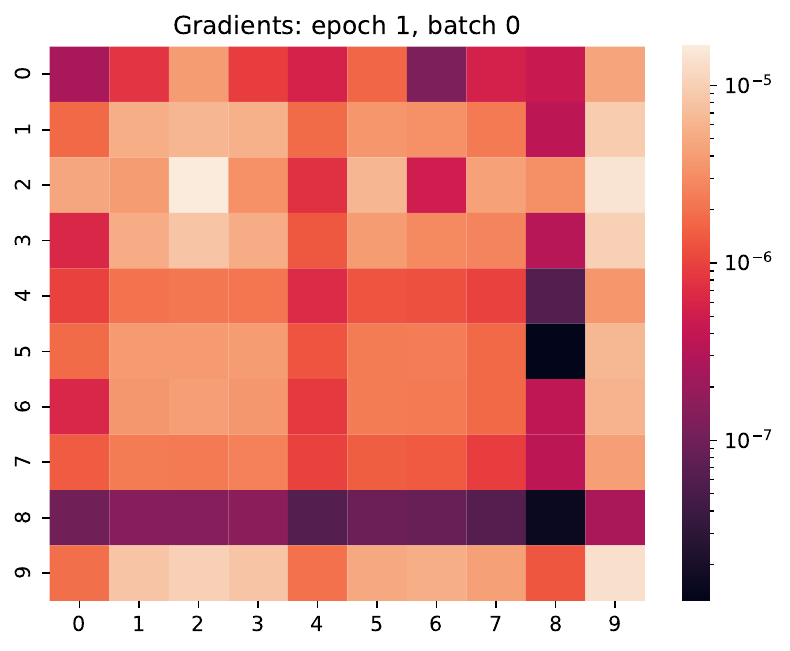}%
	\hfill
	\includegraphics[width = 0.3\linewidth]{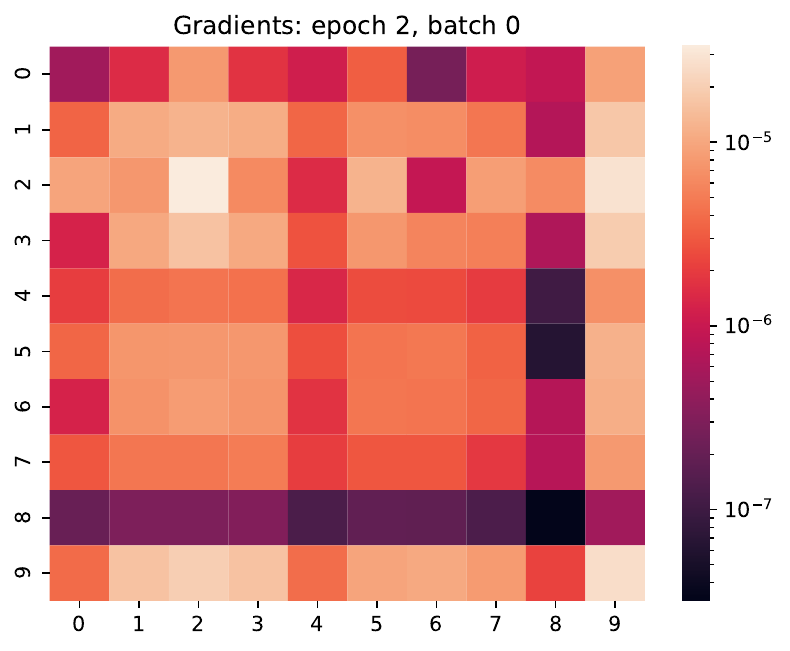}%
	\\
	\includegraphics[width = 0.3\linewidth]{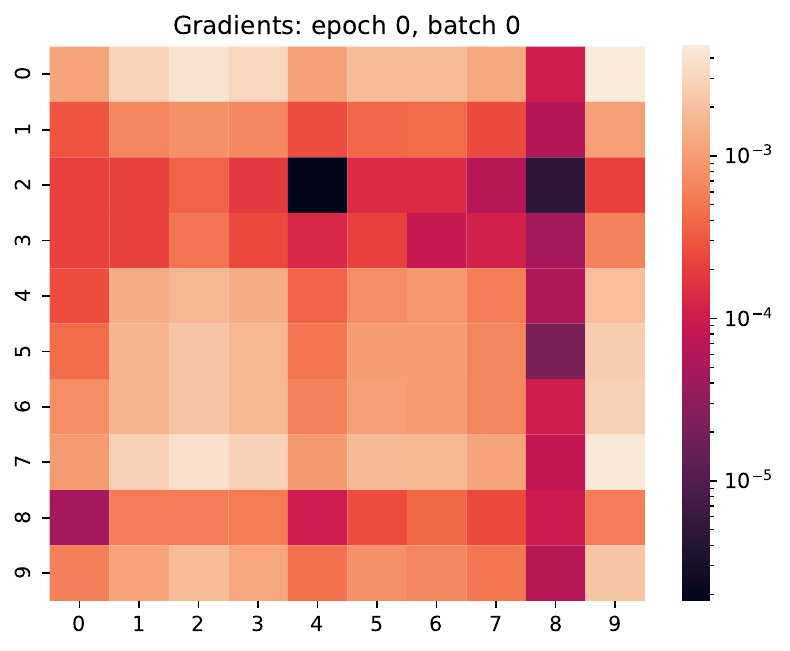}%
	\hfill
	\includegraphics[width = 0.3\linewidth]{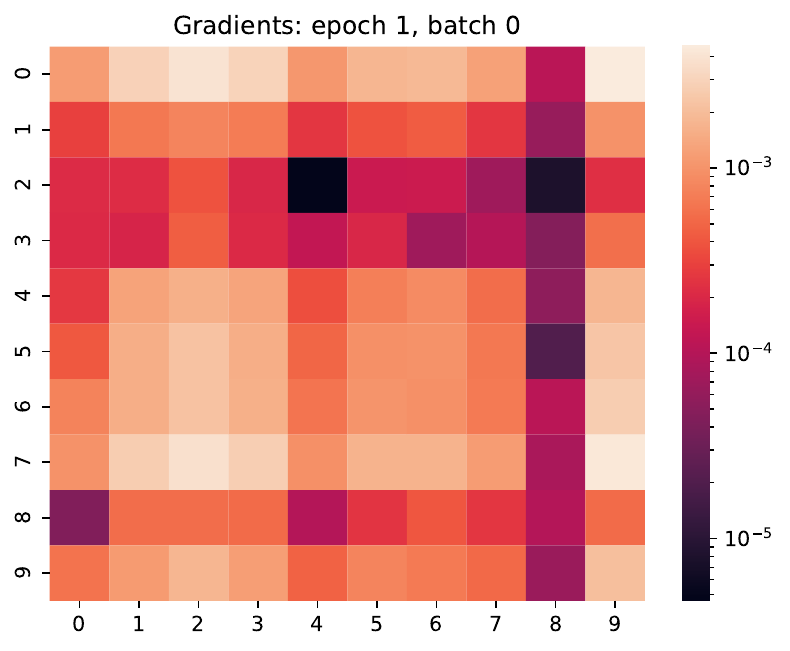}%
	\hfill
	\includegraphics[width = 0.3\linewidth]{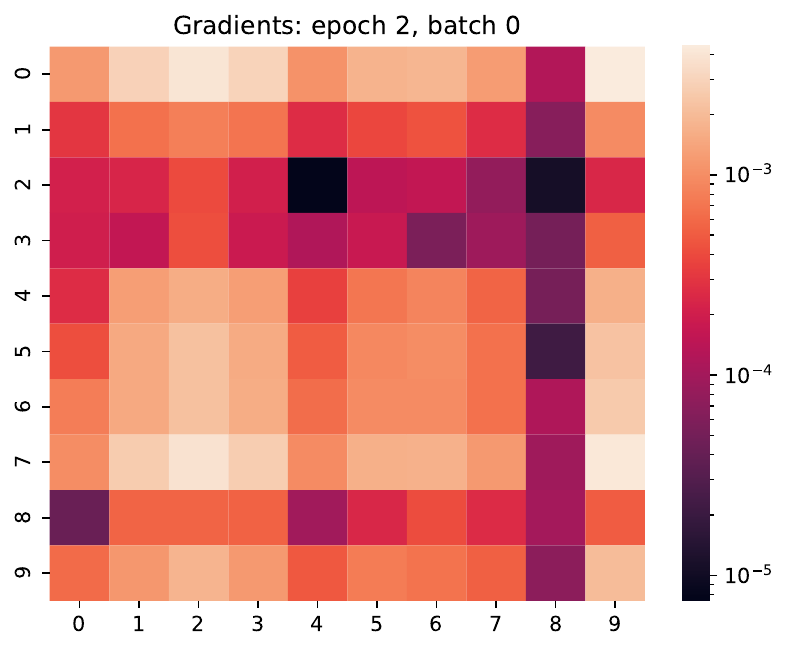}
	\caption{%
		Heatmaps showing the entries of the (Euclidean) gradient of the weight matrices of a residual layer, for the first three iterations after insertion.
		The weight matrices are newly inserted into a ResNet following the initialization described in \eqref{eq:ResNet:initialization}.
		Top row: inner weight matrix~$W_1$.
		Bottom row: outer weight matrix~$W_2$.
		Full-batch gradient descent (GD) is used as optimizer.
		The top left heatmap is white, since the gradient is zero as expected by \eqref{eq:ResNet:initialization:gradients:1}.
	}
	\label{figure:heatmaps-of-grads-of-new-weight-res}
\end{figure}

This initialization does not cause symmetry problems, as can be observed in \cref{figure:heatmaps-of-grads-of-new-weight-cnn} for CNNs, \cref{figure:heatmaps-of-grads-of-new-weight} for FNNs and \cref{figure:heatmaps-of-grads-of-new-weight-res} for ResNets.

Let us remark that while we use the initialization proposed in \citet{ChenGoodfellowShlens:2015:1} for FNNs (and CNNs), we use it to grow new layers during instead of after training (of the baseline network), which is proposed in the original paper.

\subsection{Layer Placement}
\label{subsection:layer-placement}

\begin{figure}[t]
	\centering
	\begin{subfigure}[b]{\linewidth}
		\centering
		\includegraphics[width = 0.62\linewidth]{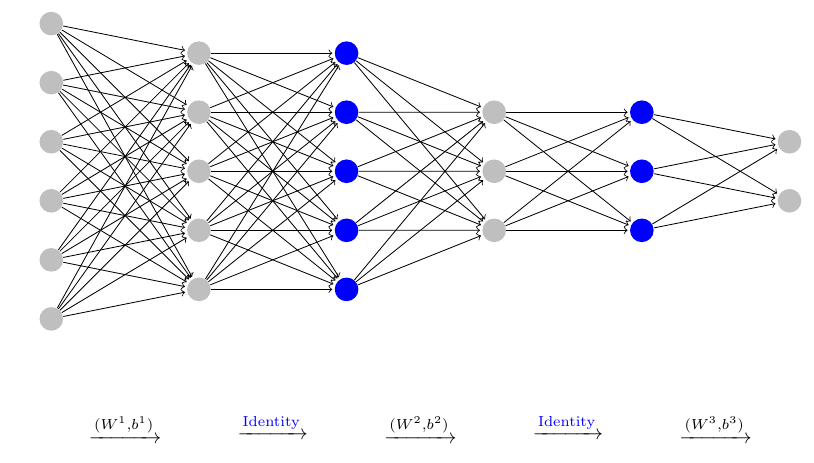}
		\caption{Feedforward Neural Network.}
		\label{figure:ReLU-FNN-positions-of-new-layers}
	\end{subfigure}
	\begin{subfigure}[b]{\linewidth}
		\centering
		\includegraphics[width = \linewidth]{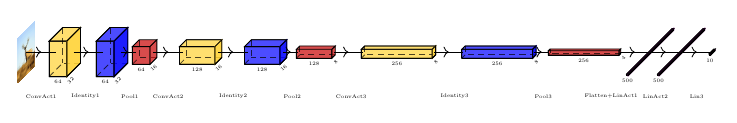}
		\caption{Convolutional Neural Network.}
		\label{figure:CNN-positions-of-new-layers}
	\end{subfigure}
	\caption{%
		Possible locations for layer insertion (blue).
	}
	\label{figure:FNN-CNN-positions-of-new-layers}
\end{figure}

The above choices for the initialization of new layers for ResNets, \cf \eqref{eq:ResNet:initialization}, \relu-FNNs, \cf \eqref{eq:FNN:initialization}, and CNNs, \cf \eqref{eq:CNN:initialization}, allow one or several new layers to be inserted.
New layers can be inserted anywhere in the network for ResNets, or anywhere after the first hidden layer in case of \relu-FNNs, or after any convolutional layer for CNNs; see \cref{figure:FNN-CNN-positions-of-new-layers} for an illustration.
In this section, we describe the SensLI procedure to find the position for a newly inserted layer to be most effective.
Insertion may happen at any point during training.
This procedure can be repeated, in order to gradually enrich a network, or used to insert multiple layers at once.

To find the best position for a new layer, we develop a notion of merit for its insertion at a particular location.
This notion borrows ideas from sensitivity analysis in constrained optimization.
To simplify notation, we write the baseline network training problem, \eqref{eq:classical-training}, in the form
\begin{equation}
	\label{eq:training-original-problem}
	\text{Minimize}
	\quad
	\fold(\thetaold)
	.
\end{equation}
For the sake of convenience of the subsequent discussion, we treat the parameters for now as column vectors, although in reality they consist of matrices and vectors.

The key point here is that we can consider the baseline problem, \eqref{eq:training-original-problem}, as a constrained version of the training problem with a new layer added.
The constrained training problem for this extended network has the form
\begin{equation}
	\label{eq:training-extended-problem}
	\begin{aligned}
		\text{Minimize}
		\quad
		&
		\fextended(\thetaextended)
		\\
		\text{\st}
		\quad
		&
		M \thetaextended - m
		=
		0
		.
	\end{aligned}
\end{equation}

The constraint in \eqref{eq:training-extended-problem} represents the initialization conditions for the new parameters, \ie, \eqref{eq:ResNet:initialization} in case of a ResNet and \eqref{eq:FNN:initialization} for a \relu-FNN.
The constraint matrix is of the form $M = [0, \id]$ so that when applied to $\thetaextended = [\thetaold, \thetanew]$, it affects only the parameters of the new layer and the vector~$m$ contains the (vectorized versions) of the initializations of all new trainable parameters (which were added in the full extension) of the original network.

The constraints in \eqref{eq:training-extended-problem} allow the baseline network to be transparently embedded into its extended counterpart.
Running a training algorithm on the baseline network, \eqref{eq:training-original-problem}, can be viewed as running the same algorithm on the extended network, with the additional weights constrained to their respective values.
With the constraints in \eqref{eq:training-extended-problem} satisfied, we have $\fextended(\thetaextended) = \fold(\thetaold)$.

Although we never actually perform any training on the constrained problem, \eqref{eq:training-extended-problem}, techniques from sensitivity analysis in constrained optimization allow us to estimate the first-order change in the value of the objective/loss~$\fextended$ \wrt a relaxation of the constraints.
This relaxation is expressed by means of
\begin{equation}
	\label{eq:training-extended-problem-perturbed}
	\begin{aligned}
		\text{Minimize}
		\quad
		&
		\fextended(\thetaextended)
		\\
		\text{\st}
		\quad
		&
		M \thetaextended - m
		=
		\perturbation
	\end{aligned}
\end{equation}
with some perturbation vector~$\perturbation$.
Classical sensitivity analysis in nonlinear programming \cite{Fiacco:1983:1} evaluates the dependence of the objective value in \eqref{eq:training-extended-problem-perturbed} on~$\perturbation$.
This is based on the implicit function theorem and typically done at a local minimizer of a nominal or reference problem ($\perturbation = 0$).
In our setting, however, we need to be able to apply the technique at any point during the training.
Since the training algorithm runs on the baseline network, it does not employ the additional, virtual parameters~$\thetanew$.
These parameters only get inserted when it is time to extend the network, and they are chosen to satisfy the constraint $M \thetaextended - m = 0$ at insertion time exactly.
However, $\thetaold$ is generally not a minimizer for the baseline model, \eqref{eq:training-original-problem}.
Therefore, a residual $r = \nabla \fold(\thetaold)$ remains at the time training on the baseline model is suspended.

In order to apply sensitivity analysis,  the current point $\thetaold$ is treated as a local solution, or at least a stationary point, of a problem with modified objective $\fold(\thetaold) - r^\transp \thetaold = \fextended(\thetaextended) - r^\transp \thetaold$.
With the modified objective, $\thetaextended$ becomes a stationary point of the extended problem, \eqref{eq:training-extended-problem-perturbed}, at $\perturbation = 0$.
Since the constraints are linear and the constraint Jacobian~$M$ has linearly independent rows, the linear independence constraint qualification (LICQ) holds.
Consequently, $\thetaextended$ is a KKT point with uniquely defined Lagrange multiplier vector $\lambda$ satisfying
\begin{subequations}
	\label{eq:KKT-conditions}
	\begin{align}
		\nabla \fextended(\thetaextended)
		-
		\begin{bsmallmatrix}
			r
			\\
			0
		\end{bsmallmatrix}
		+
		M^\transp \lambda
		&
		=
		0
		,
		\\
		M \thetaextended - m
		&
		=
		0
		.
	\end{align}
\end{subequations}
Taking into account the partitioning $\thetaextended = [\thetaold, \thetanew]$ and the structure of $M = [0, \id]$, we can write \eqref{eq:KKT-conditions} as
\begin{subequations}
	\label{eq:KKT-conditions:detailed}
	\begin{align}
		\nabla_{\thetaold} \fextended(\thetaextended) - r
		&
		=
		0
		,
		\label{eq:KKT-conditions:detailed:1}
		\\
		\nabla_{\thetanew} \fextended(\thetaextended) + \lambda
		&
		=
		0
		,
		\label{eq:KKT-conditions:detailed:2}
		\\
		\thetanew - m
		&
		=
		0
		.
		\label{eq:KKT-conditions:detailed:3}
	\end{align}
\end{subequations}
Results from sensitivity analysis, \eg, \citet[Theorem~3.2.2]{Fiacco:1983:1} or \citet[Theorem~2.24]{ItoKunisch:2008:1} now imply that, under suitable conditions, $\lambda^\transp \perturbation$ is the directional derivative of the value of the modified objective $\thetaextended \mapsto \fextended(\thetaextended) - r^\transp \thetaold$ in the direction~$\perturbation$.
Therefore, a relaxation of the constraint in the direction of~$\perturbation$, see \eqref{eq:training-extended-problem-perturbed}, results in a predicted first-order change in the value of the objective of $\lambda^\transp \perturbation = - \nabla_{\thetanew} \fextended(\thetaextended)^\transp \perturbation$, where we used \eqref{eq:KKT-conditions:detailed:2}.
Consequently, the (Euclidean) norm
\begin{equation}
	\label{eq:notion-of-merit}
	\norm{\nabla_{\thetanew} \fextended(\thetaextended)}
\end{equation}
provides a notion of merit of inserting the layer with parameters~$\thetanew$.

For simplicity, the description above was referring to a single layer and the merit of its insertion into the network.
In fact, the analysis does not change when several layers are considered to be added simultaneously.
Put simply, $\thetanew$ now contains the trainable parameters of each potential layer as subvectors, and the same structure will be inherited by the Lagrange multiplier~$\lambda = - \nabla_{\thetanew} \fextended(\thetaextended)$.
We can therefore evaluate the norm, \eqref{eq:notion-of-merit}, separately for each chunk of parameters pertaining to a particular layer, and compare them.
In our implementation, we are using the Frobenius norm of the partial gradient \wrt the weight matrix~$W$,
\begin{equation}
	\label{eq:notion-of-merit-per-layer:ReLU-FNN}
	\norm{\nabla_W \fextended(\thetaextended)}_F^2
\end{equation}
as our final notion of merit of inserting a layer in case of a \relu-FNN.
The impact of the bias vector is disregarded.
In case of a ResNet, we use \eqref{eq:notion-of-merit-per-layer:ReLU-FNN} with the outer weight matrix~$W_2$ instead of~$W$.
The reason we can disregard the impact of $W_1$ is that, according to \eqref{eq:ResNet:initialization:gradients:1}, $\nabla_{W_1} \fextended(\thetaextended)$ is initially equal to zero.

In CNNs, $\nabla_K \fextended(\thetaextended)$ is a higher-dimensional object than the gradient in an FNN.
It is therefore sensible to choose a norm suited to the structure of the kernel.
We compare the behavior of different norms in \cref{section:sensitivity-norms} and employ the operator norm, \eqref{eq:notion-of-merit-per-layer:CNN}, in what follows.
We select the layer maximizing the merit as the single layer to be inserted at this point in training.
We remark that a scaling factor $\frac{1}{h_k^2}$ may be used to make the notion of merit comparable across different layers in terms of computational complexity of the layer in case for FNNs.

The evaluation of the Frobenius norm, \eqref{eq:notion-of-merit-per-layer:ReLU-FNN}, is simple and numerically inexpensive.
To obtain the gradients, we suspend training and temporarily build a fully extended network with layers added at all potential positions.
We populate the weights and biases by copying the current value of $\thetaold$ into the respective positions and initializing the newly added weights and biases as described in \cref{subsection:initialization}.
We then evaluate $\fextended(\thetaextended)$ once and simultaneously its gradients \wrt all parameters by a single forward-backward pass, without updating any of the weights.
This corresponds to a single full-batch gradient descent step with zero learning rate.
The quantity in \eqref{eq:notion-of-merit-per-layer:ReLU-FNN} can then be easily evaluated layer by layer.

In case of a mini-batch training algorithm, $\nabla_W \fextended(\thetaextended)$ is not available at once.
We therefore perform an entire epoch of mini-batch SGD steps with zero learning rate to get access to $\nabla_W \fextended(\thetaextended)$ before evaluating its norm, \eqref{eq:notion-of-merit-per-layer:ReLU-FNN}.
It would be possible to trade accuracy for computational efficiency by computing the sensitivities on a (large) mini-batch instead.
Having determined the largest of these norms across all potential new layers, we discard the fully extended network, insert the selected layer into the baseline model, initialize the new weights and biases as described in \cref{subsection:initialization}, and resume training.

\subsection{When to Insert a Layer?}
\label{subsection:when-to-insert}

The optimal timing for suspending training to insert a new layer remains an open question without a satisfying answer.
On the one hand, training the smaller network for as long as possible saves resources.
On the other hand, training it for too long can be inefficient and may prevent the extended network from fully leveraging the added layer.
In our numerical experiments shown in \cref{figure:when-to-insert-layers}, we find that the time when to insert a new layer can have a significant impact on the effectiveness of the overall approach.
To the best of our knowledge, currently existing methods to decide the insertion time during training are heuristic.
Similarly, we propose to evaluate the merit of inserting a new layer at regular intervals during training, and to insert the layer with the highest sensitivity when
\begin{equation}
	\label{eq:when-to-insert}
	\frac{\norm{\nabla_{\thetanew} \fextended(\thetaextended)}^2}{\frac{1}{\# \cW} \sum_{W \in \cW} \norm{\nabla_W \fextended(\thetaextended)}^2}
	\ge
	\tau
\end{equation}
holds with $\tau \ge 1$.
Here, $\cW$ is the set containing all weight matrices or kernels already present in the baseline model.
Hence a layer is selected for insertion when the sensitivity of this layer relative to a suitable norm of the gradient of all parameters in the network is sufficiently large.
In \cref{figure:performance-of-layer-insertion-cnn}, we observe that already for $\tau = 1$ the heuristic in \eqref{eq:when-to-insert} actually accepts and rejects layer insertions, \cf \cref{table:parameters-cnn}.

\subsection{Statement of the Algorithm}
\label{subsection:statement-of-the-algorithm}

We summarize the steps of the SensLI algorithm in \cref{alg:sensli}.
\begin{algorithm}[ht]
	\caption{Sensitivity-based Layer Insertion (SensLI)}
	\label{alg:sensli}
	\begin{algorithmic}[1]
		\State Start with a baseline architecture $\networkold$ (either an FNN, ResNet, or CNN as described in \cref{subsection:notation} and \cref{subsection:notation-cnn}) and initial parameters $\theta_0$.
		\State Train on the baseline network for $K_0$~epochs, starting from~$\theta_0$  (with a chosen optimizer): $\theta_{\text{curr}} \coloneqq \text{Train}(\networkold,\theta_0,K_0)$.
		\While{termination criterion not met}
		\State Fully extend the current network by inserting identity layers at all possible positions: $\networkextended \coloneqq \text{Extend}(\networkold)$.
		\State Initialize the fully-extended network with $\thetaextended$ as described in \cref{subsection:initialization}: $\thetaextended \coloneqq \text{InitExtended}(\networkextended, \theta_{\text{curr}})$.
		\State Compute sensitivities \wrt the new weight matrices over the whole training data by performing a backward pass for every training data point as described in \cref{subsection:layer-placement}: $\lambda \coloneqq \text{SensComp}(\networkextended, \thetaextended)$.
		\State Select whether and where to insert a layer by comparing the norms of the sensitivities and
		checking the threshold in \eqref{eq:when-to-insert}: $\networkold, \theta_0 \coloneqq \text{SelectNewNetwork}(\networkold,\theta_{\text{curr}},\lambda)$.
		\State Train on the current network for $K_{\text{curr}}$~epochs, starting from~$\theta_0$ (with a chosen optimizer): $\theta_{\text{curr}} \coloneqq \text{Train}(\networkold,\theta_0,K_{\text{curr}})$.
		\EndWhile
		\State \Return extended network $\networkold$ with parameters $\theta_{\text{curr}}$.
	\end{algorithmic}
\end{algorithm}

\FloatBarrier
\section{Numerical Results}
\label{section:numerical-results}

\begin{wrapfigure}{r}{0.3\linewidth}
	\centering
	\includegraphics[width = \linewidth]{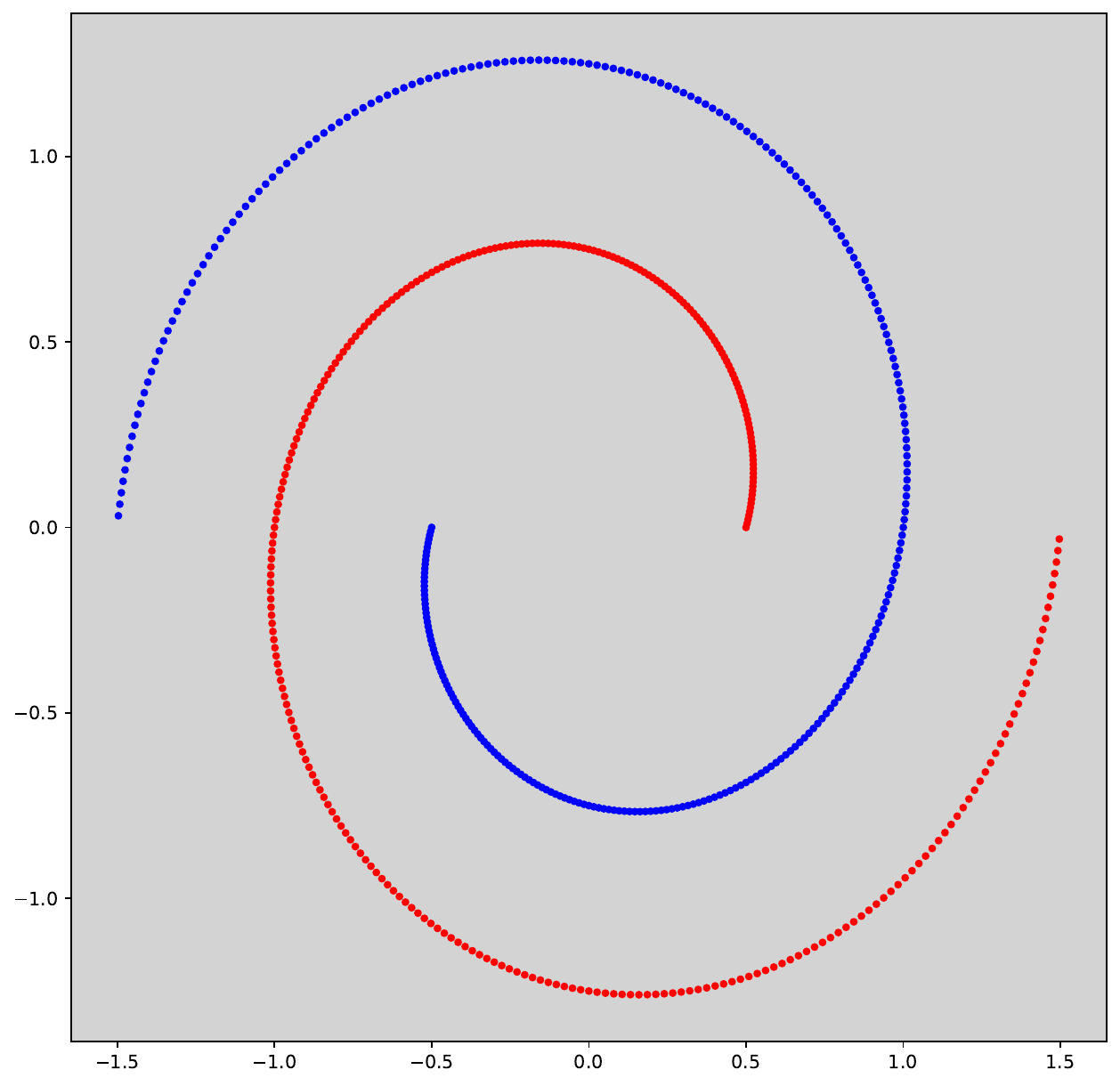}
	\caption{Illustration of the spiral data set.}
	\label{figure:illustration-spiral-dataset}
\end{wrapfigure}
Our implementation is available on \github\footnote{\url{https://github.com/mathemml/SensLI}} and is based on \pytorch by \citet{PaszkeGrossMassaLererBradburyChananKilleenLinGimelsheinAntigaDesmaisonKoepfYangDeVitoRaisonTejaniChilamkurthySteinerFangBaiChintala:2019:1}.

We consider a spiral data set for experiments with fully connected networks, \cf \cref{figure:illustration-spiral-dataset}.
The data set consists of 600~data points, where each data point is composed of a two-dimensional feature vector $x_i \in \R^2$ and a label that indicates whether it belongs to the red or blue spiral, \ie, $y_i \in \set[auto]{\begin{psmallmatrix} 0 \\ 1 \end{psmallmatrix}, \begin{psmallmatrix} 1 \\ 0 \end{psmallmatrix}} \subset \R^2$.
For training, we use 450~data points, while the remaining 150~data points form the test set.
In numerical experiments with CNNs we employ the CIFAR-10 data set \citep{Krizhevsky:2009:1}.
Here we use the standard data augmentation, a train-test split of \num{50000} and \num{10000} images, respectively.
We do not require a validation set since no hyperparameter search is performed.
In fact, our depth-adaptive layer insertion method can be viewed as an automated hyperparameter search for the depth of the network.

Detailed setup information for all experiments can be found in \cref{section:hyperparameters}.
Full-batch gradient descent allows for the best interpretation of results, which is why we employ it in several experiments.
Although mini-batch SGD (with momentum) introduces additional uncertainty, it can also deliver more competitive results, especially in applications with larger data sets.
Therefore, we also show experiments using mini-batches.
In all plots, layer insertion is indicated by a vertical dotted line.

\subsection{Comparison to Fixed-Architecture Training}
\label{subsection:comparison-to-fixed-architecture-training}

In this subsection we demonstrate that SensLI can be beneficial compared to fixed-architecture training.
We first present experiments with only a single layer insertion.
Subsequently, we will consider repeated layer insertions as in \cref{figure:performance-of-layer-insertion-cnn}.

\subsubsection{Insertion of a Single Layer}

We compare SensLI applied to FNN, ResNet and CNN architectures (FNN~LI, ResNet~LI, CNN~LI) to the respective baseline (FNN1, ResNet1, CNN1) and extended architectures (FNN2, ResNet2, CNN2).
Here, the number of network parameters of the baseline architectures coincides with those in the first epochs of SensLI, while the extended architecture has the same number of parameters as the SensLI architecture after layer insertion, \cf \cref{table:parameters-fb}.

\begin{table}[h]
	\centering
	\caption{Number of network parameters during training.}
	\label{table:parameters-fb}
	\resizebox{0.7\linewidth}{!}{
		\begin{tabular}{rrrrrr}
			\toprule
			Epochs    & FNN LI & ResNet LI & & Epochs  & CNN LI
			\\
			\cmidrule{1-3}
			\cmidrule{5-6}
			0--449    & 27     & 33        & & 0--49   & \num{2674816}
			\\
			450--end  & 57     & 54        & & 50--end & \num{2711744}
			\\
			\bottomrule
		\end{tabular}
	}
\end{table}

As a theoretical comparison, we provide approximate counts of floating point operations (FLOPs) for the different architectures, \cf \cref{table:flops-comparison}.
Specifically, we report on the number of FLOPs needed for SensLI evaluation to get a better understanding of the relative effort required.
Note that for the larger CNN architecture the SensLI evaluation is more costly due to the higher number of parameters in the extended layers.
Simultaneously, we observe a more substantial FLOP reduction from CNN2 to CNN~LI than in the smaller FNN and ResNet setups.

\begin{table}[h]
	\centering
	\caption{Comparison of FLOPs per training sample for \cref{figure:comparison-with-fixed-architecture}.}
	\label{table:flops-comparison}
	\begin{tabular}{lrl}
		\toprule
		\textbf{Architecture} & \textbf{FLOPs}          & \textbf{thereof FLOPs for SensLI evaluation}
		\\
		\cmidrule{1-3}
		FNN1                  & \num{222000}               &
		\\
		FNN2                  & \num{499500}               &
		\\
		FNN LI                & \num{432270}               & 270, approx. \SI{0.06}{\percent}
		\\
		ResNet1               & \num{333000}               &
		\\
		ResNet2               & \num{532000}               &
		\\
		ResNet LI             & \num{484488}               & 288, approx. \SI{0.05}{\percent}
		\\
		CNN1                  & \num{25196544000}        &
		\\
		CNN2                  & \num{47845785600}        &
		\\
		CNN LI                & \num{37452607488}        & \num{931442688}, approx. \SI{2.5}{\percent}
		\\
		\bottomrule
	\end{tabular}
\end{table}

\begin{figure}[tbp]
	\centering
	\begin{subfigure}{0.57\linewidth}
		\includegraphics[width = 0.995\linewidth, right]{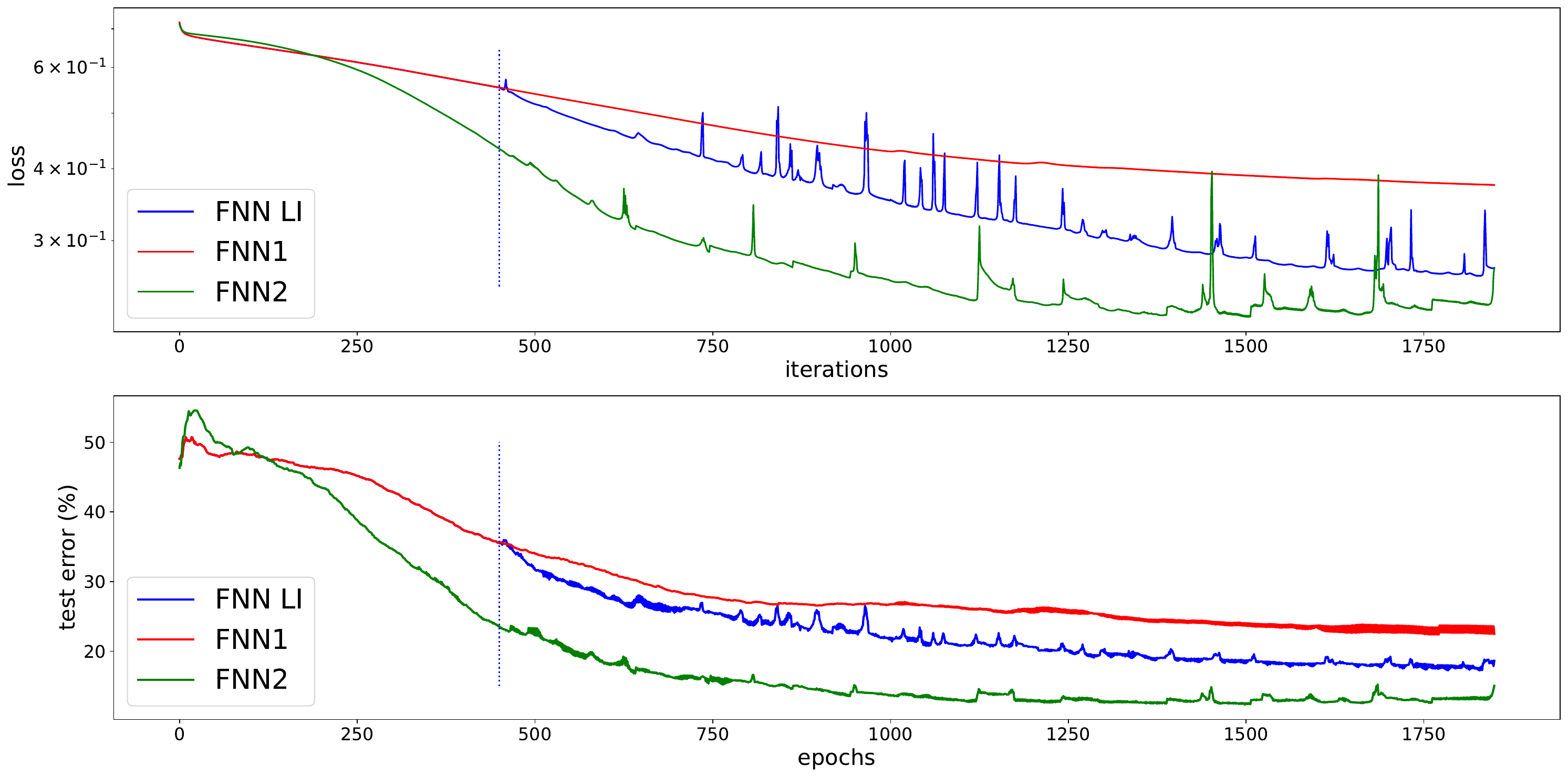}
		\caption{FNN, full-batch GD.}
		\label{figure:comparison-with-fixed-architecture-fnns}
	\end{subfigure}

	\begin{subfigure}{0.57\linewidth}
		\includegraphics[width = \linewidth, right]{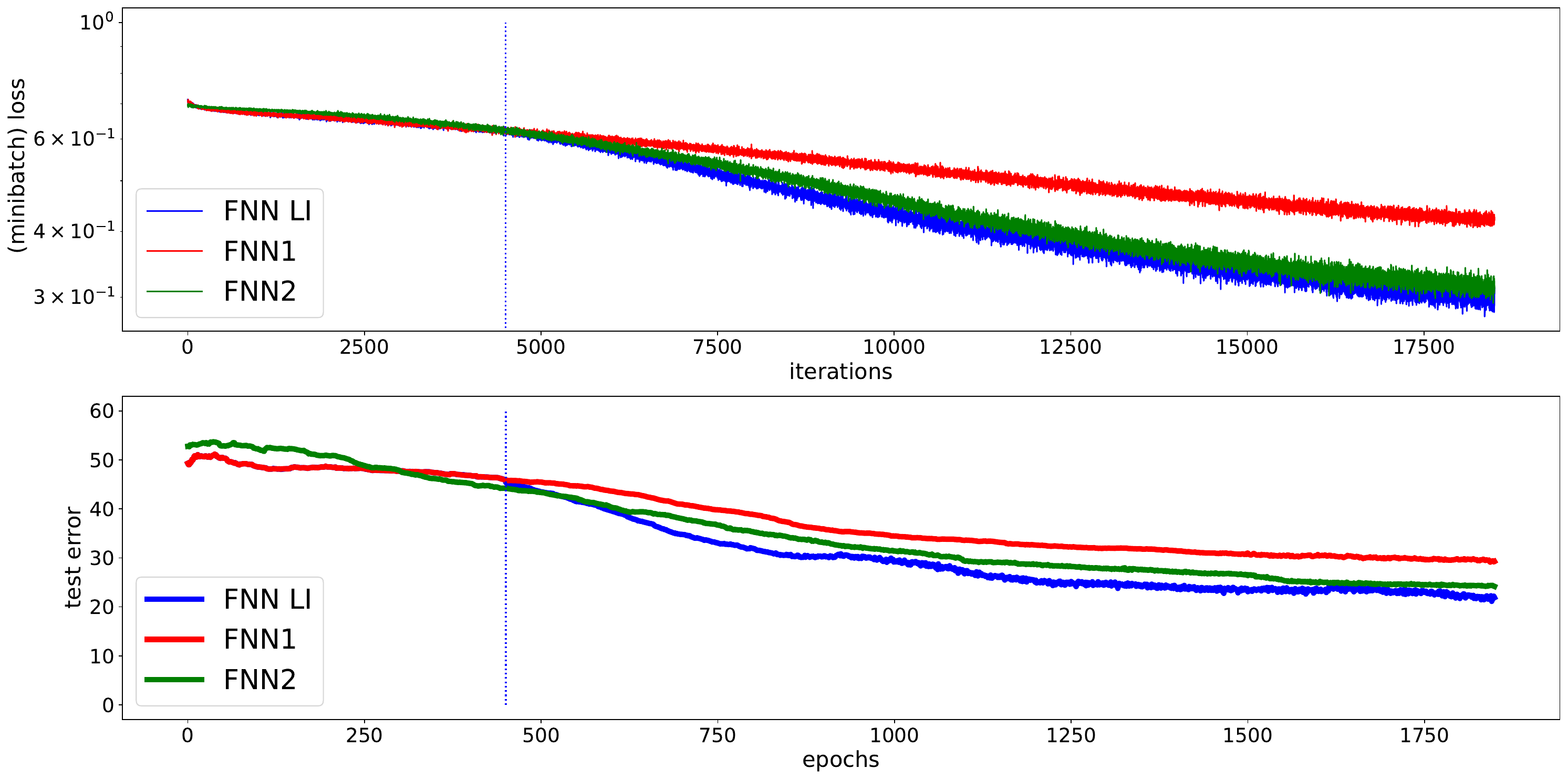}
		\caption{FNN, mini-batch SGD.}
		\label{figure:comparison-with-fixed-architecture-mb}
	\end{subfigure}

	\begin{subfigure}{0.57\linewidth}
		\includegraphics[width = 0.995\linewidth, right]{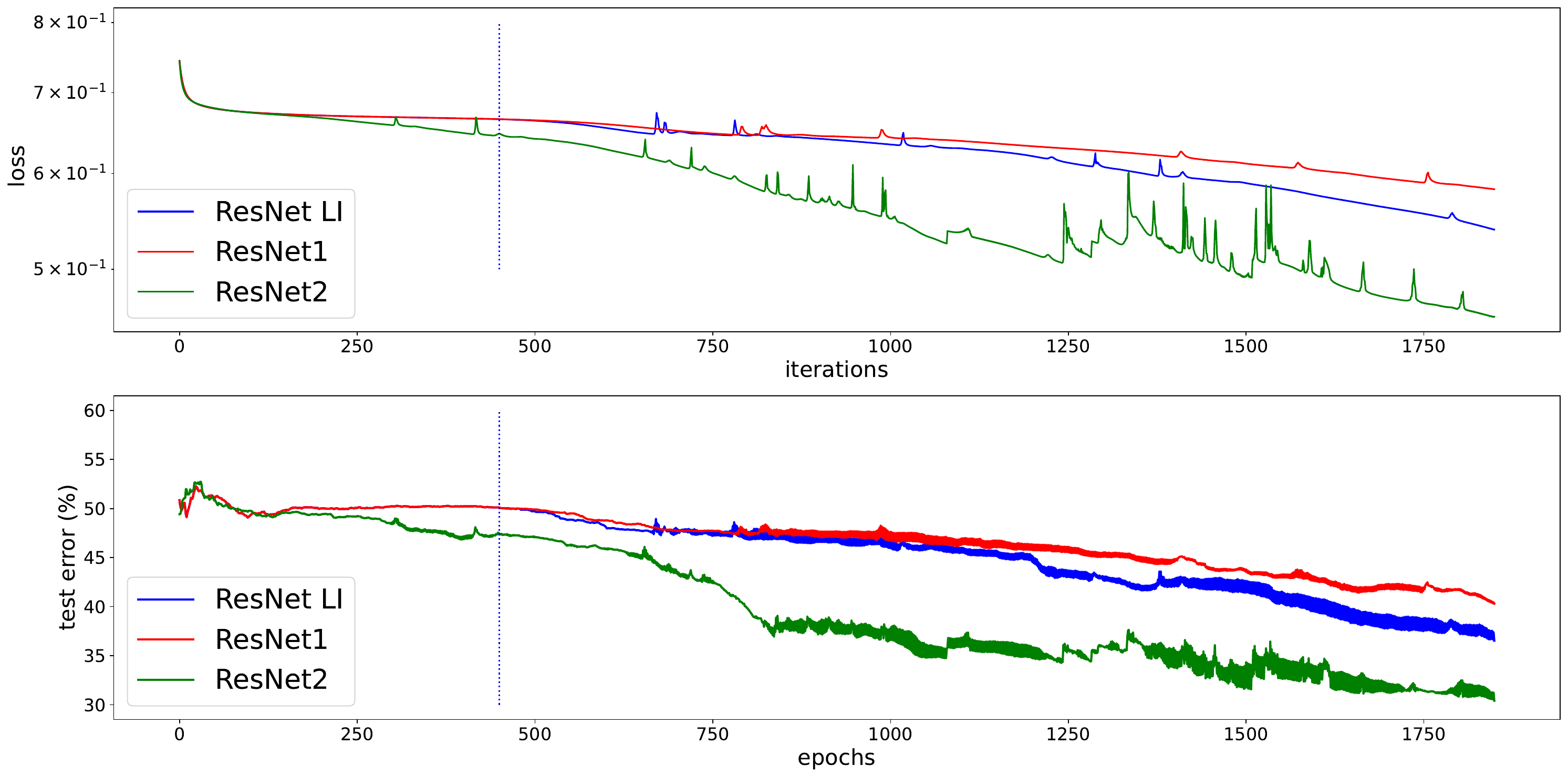}
		\caption{ResNet, full-batch GD.}
		\label{figure:comparison-with-fixed-architecture-resnets}
	\end{subfigure}

	\begin{subfigure}{0.57\linewidth}
		\includegraphics[width = 0.97\linewidth, right]{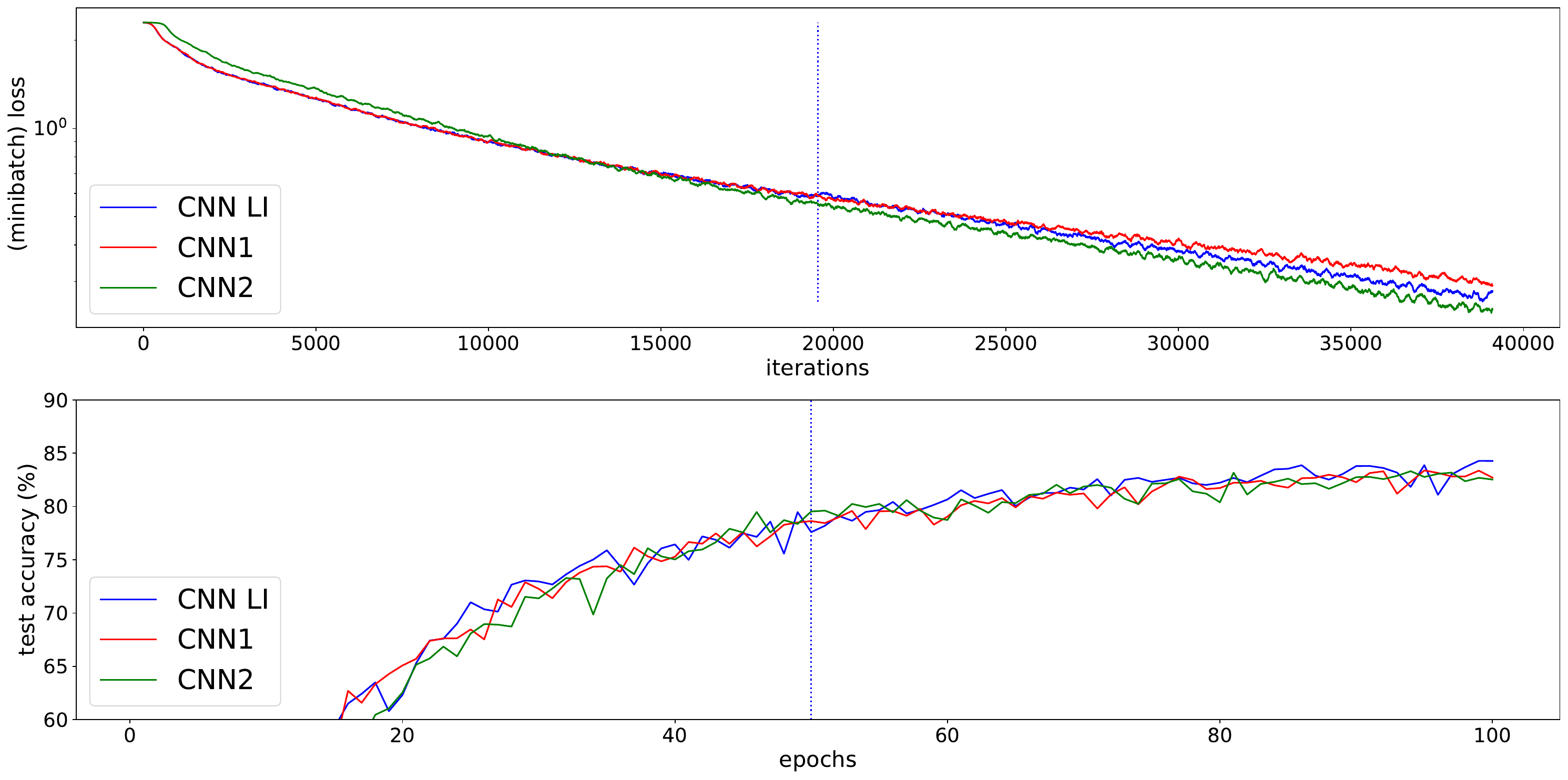}
		\caption{CNN, mini-batch SGD with momentum.}
		\label{figure:comparison-with-fixed-architecture-cnn}
	\end{subfigure}
	\caption{%
		Comparison of layer insertion and fixed-architecture training for FNNs with full-batch (\subref{figure:comparison-with-fixed-architecture-fnns}) and mini-batch SGD (\subref{figure:comparison-with-fixed-architecture-mb}), ResNets with full-batch (\subref{figure:comparison-with-fixed-architecture-resnets}) and CNNs with mini-batch SGD.
		We show the loss (top) and test error (bottom) averaged over 30 (FNN), 40 (ResNet) and 10 (CNN) runs, respectively.
		These experiments are included in the \github repository as Exp2--Exp5.
	}
	\label{figure:comparison-with-fixed-architecture}
\end{figure}

Due to the influence of random initialization, we average over multiple training runs in \cref{figure:comparison-with-fixed-architecture}.
We fix the iteration at which a layer is inserted and also the learning rate for comparability of multiple runs.
Restoring the random seed, we use identical initializations for the baseline and SensLI architectures.
Consequently, the loss and error histories agree until layer insertion, except for the noise introduced by mini-batch selection.

\FloatBarrier

For all architectures, we see that SensLI leads to an accelerated loss decay compared to fixed-architecture training on the \emph{baseline} network.
This indicates that the inserted layer is actively contributing to the learning process.
The effect is most pronounced for FNNs, especially in combination with mini-batch SGD, where SensLI also outperforms training on the extended network FNN2.
For ResNets and CNNs, the advantage of layer insertion is less pronounced.
A possible explanation is that the baseline architectures ResNet1 and CNN1 are richer from the beginning than the baseline architecture FNN1,
and that both the absolute and relative increase in the number of parameters upon layer insertion are smaller.

When SensLI is compared to fixed-architecture training on the \emph{extended} networks, the loss decay is only accelerated for FNNs with mini-batch SGD.
However, in all cases, SensLI is numerically cheaper, \cf \cref{table:flops-comparison}.
Let us further remark that we fixed the insertion epoch for comparability, which may limit the potential of SensLI since the best insertion epoch depends on the initialization.
Hence, in this setup, we cannot expect SensLI to be equally effective for all training runs that are being averaged over.
Finally, since the number of parameters in the extended CNN2 model is only \SI{1.4}{\percent} larger compared to the baseline CNN1 model (\cref{table:parameters-fb}),
it is not unexpected to see only minor differences in accuracy.

\begin{figure}[hb]
	\centering
	\begin{subfigure}{\linewidth}
		\includegraphics[width = 0.99\linewidth, right]{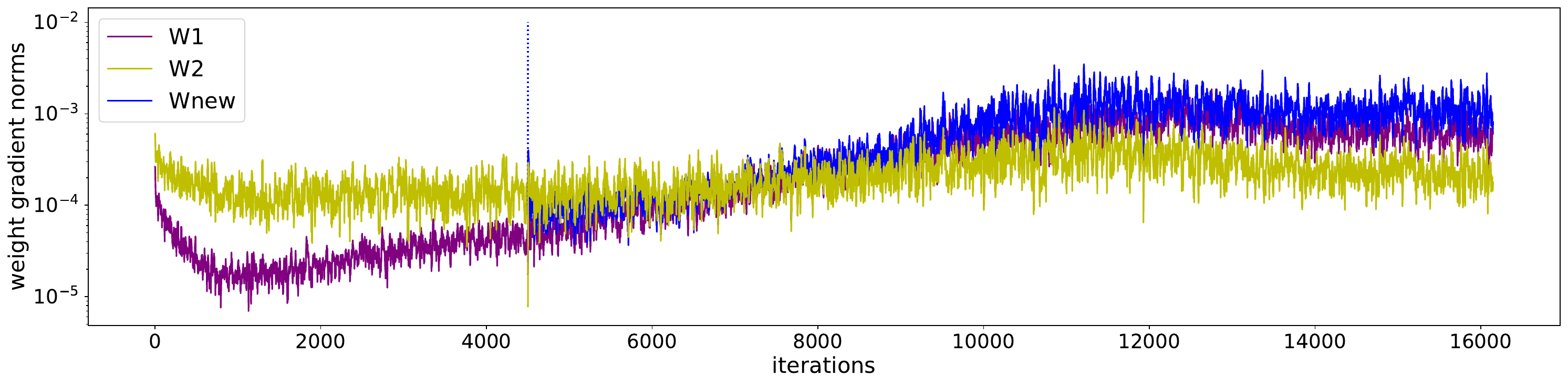}
		\caption{FNN, mini-batch SGD.}
		\label{figure:LI-layerwise-grads-mb}
	\end{subfigure}
	\hfill
	\begin{subfigure}{\linewidth}
		\includegraphics[width = \linewidth, right]{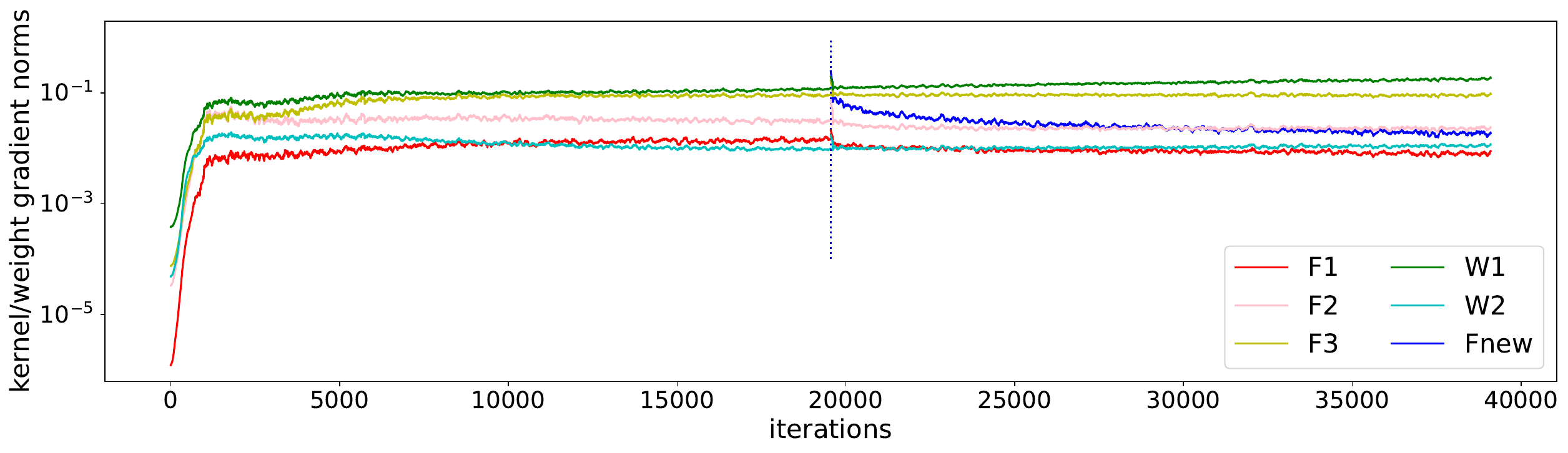}
		\caption{CNN, mini-batch SGD with momentum.}
		\label{figure:LI-layerwise-grads-cnn}
	\end{subfigure}
	\caption{%
		Behavior of layerwise gradients during training for an FNN with mini-batch SGD (\subref{figure:LI-layerwise-grads-mb}) and a CNN with mini-batch SGD (\subref{figure:LI-layerwise-grads-cnn}).
		We show the Frobenius norm of the gradients of the trainable parameters in the networks over iteration count (newly inserted weight matrix in blue).
		The experiments can be found in the \github repository as Exp4 and Exp5.
	}
	\label{figure:LI-layerwise-grads}
\end{figure}

\Cref{figure:LI-layerwise-grads} shows one exemplary run using SensLI for an FNN with mini-batch SGD and a CNN with mini-batch SGD with momentum, respectively.
The newly inserted weight matrices take gradient norm values (measured in the Frobenius norm) within a comparable range of the norms of pre-existing trainable values, \ie, no vanishing or exploding gradient problems are encountered.
For the FNN architecture, the new gradient exceeds the other gradients, which agrees with the pronounced effect of layer insertion visible in \cref{figure:comparison-with-fixed-architecture-mb}.
Meanwhile, for the CNN, the new gradient is not dominating, reflected by the minor effect of layer insertion in \cref{figure:comparison-with-fixed-architecture-cnn}.

The previous experiments illustrate that a single execution of SensLI can outperform fixed-architecture training, but in some scenarios only has a slight effect.
We now investigate the training performance of SensLI with repeated layer insertions.

\subsubsection{Repeated Insertion of Layers}

In \cref{figure:performance-of-layer-insertion-cnn} we compare SensLI for a CNN architecture with training on the extended CNN from the beginning.
We can see that SensLI uses $\SI{1659}{\second}/\SI{2320}{\second} = \SI{71.5}{\percent}$ of the time that the extended CNN needs to complete 200~epochs.
As a theoretical comparison, we also calculate the approximate number of FLOPs (per training data point), summed over all epochs as follows:
The fixed-architecture training on the extended CNN uses around $\num{46996684800} \times 3$  FLOPs per training data point, while SensLI uses around $\num{33063616512} \times 3$ FLOPs per training data point, thereof $\num{914460672} \times 3$ for the layer insertion evaluation, which amounts to approximately $\SI{2.8}{\percent}$ of the total training cost.
The ratio of FLOPs per training data point ($\text{SensLI}/\text{CNN} = \SI{70}{\percent}$) does almost directly translate to total computational time.
The small deviation is most likely due to the fact that initialization and setup also require computing time.

\begin{figure}[hb]
	\centering
	\includegraphics[width = \linewidth]{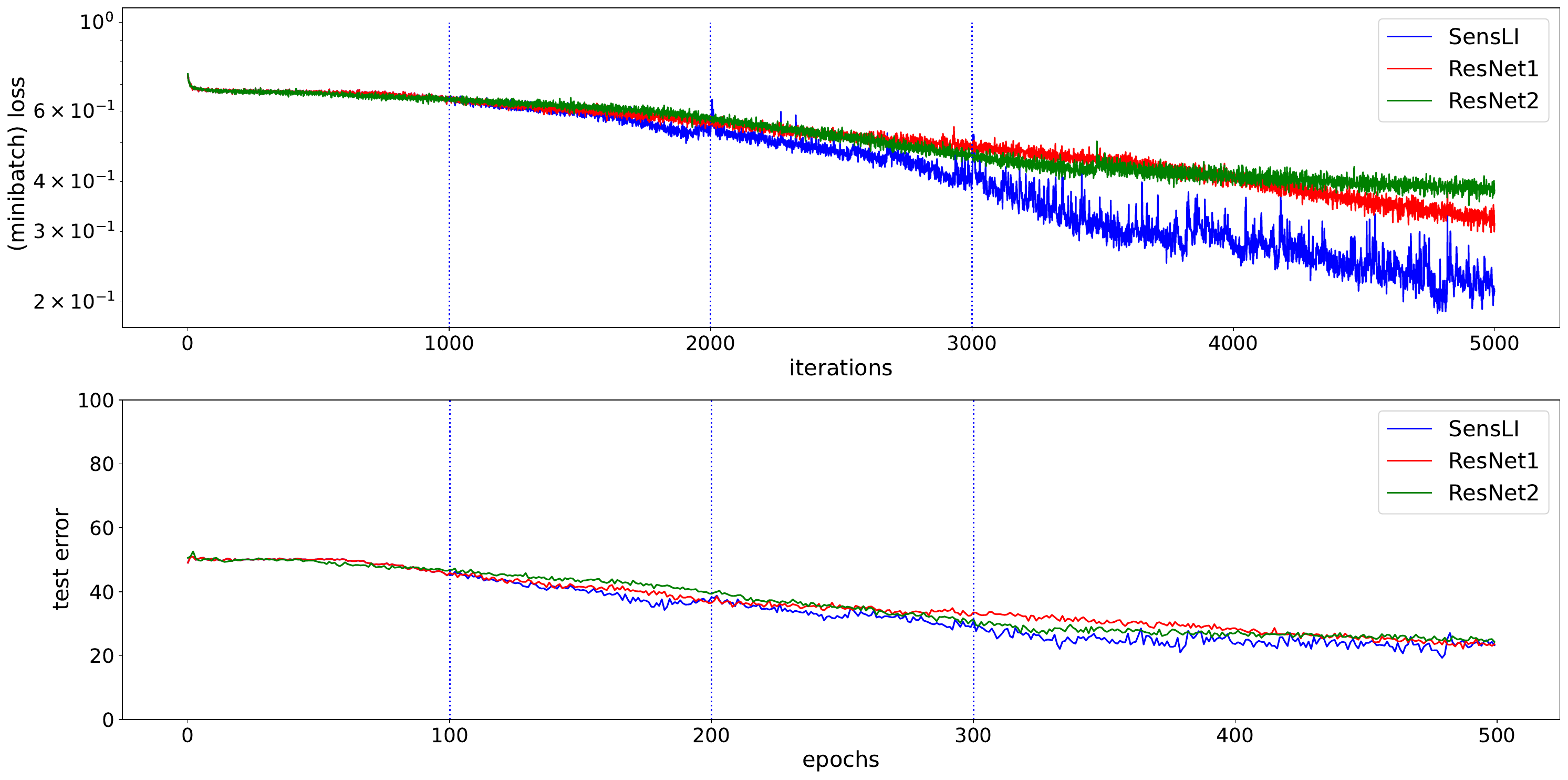}
	\caption{%
		Comparison of SensLI and fixed-architecture training for ResNets with mini-batch SGD.
		SensLI inserts a layer three times in the training process, indicated by vertical lines.
		We show the loss (top) and test error (bottom) averaged over 30 (ResNet) runs.
		The experiments can be found in the \github repository as Exp13.
	}
	\label{figure:comparison-with-fixed-architecture-resnets-3lis}
\end{figure}

As an additional experimental setup, we consider a ResNet architecture with three repetitions of layer insertion, \cf \cref{figure:comparison-with-fixed-architecture-resnets-3lis}.
The ResNet architecture is trained with mini-batch SGD and the learning rate is fixed to \num{0.01} for all runs.
After 100, 200 and 300~epochs, we let SensLI insert a new layer.
We observe that the performance of SensLI is better on average than the fixed-architecture training on the baseline ResNet1 and even outperforms the extended ResNet2 architecture.
Hence, SensLI manages to leverage the additional capacity of the inserted layers, while not getting stuck in local minima which can be observed for fixed-architecture training on ResNet2.
Comparing the behaviour of SensLI in \cref{figure:comparison-with-fixed-architecture} with \cref{figure:comparison-with-fixed-architecture-resnets-3lis} (and \cref{figure:performance-of-layer-insertion-cnn}),
we see that the performance of SensLI is more pronounced in the latter case with multiple layer insertions.
Further, SensLI takes less time to train the ResNet architecture with three layer insertions than the fixed-architecture training on the extended ResNet2 architecture, \cf \cref{table:flops-comparison-resnet-3lis}.

\begin{table}[h]
	\centering
	\caption{Comparison of FLOPs per training sample for \cref{figure:comparison-with-fixed-architecture-resnets-3lis}.}
	\label{table:flops-comparison-resnet-3lis}
	\begin{tabular}{lrl}
		\toprule
		\textbf{Architecture} & \textbf{FLOPs}          & \textbf{thereof FLOPs for SensLI evaluation}
		\\
		\cmidrule{1-3}
		ResNet1               & \num{90000}                &
		\\
		ResNet2               & \num{252000}                &
		\\
		ResNet LI             & \num{188712}                & \num{1512}, approx. \SI{0.8}{\percent}
		\\
		\bottomrule
	\end{tabular}
\end{table}

\subsection{Comparison of Inserted Layer Placement}
\label{subsection:validity-of-insertion-positioning}

In \cref{subsection:layer-placement} we propose a strategy to determine the most promising position for a new layer to be inserted, based on a notion of merit using sensitivity analysis of the objective function, which we aim to validate here.
For the sake of comparability, we use baseline architectures of limited depth, where all potential layers for insertion have the same number of parameters, \cf \cref{section:hyperparameters} Exp6 to Exp9.

We consider the proposed SensLI strategy (based on the largest merit indicator) and compare it to the exact opposite strategy (LIother) that uses the smallest indicator.
We employ the merit indicators \eqref{eq:notion-of-merit-per-layer:ReLU-FNN} with respect to $W$ or $W_2$ and \eqref{eq:notion-of-merit-per-layer:CNN} for FNN, ResNet and CNN architectures, respectively.

\begin{figure}[tb]
	\centering
	\begin{subfigure}[c]{\linewidth}
		\includegraphics[width = \linewidth, right]{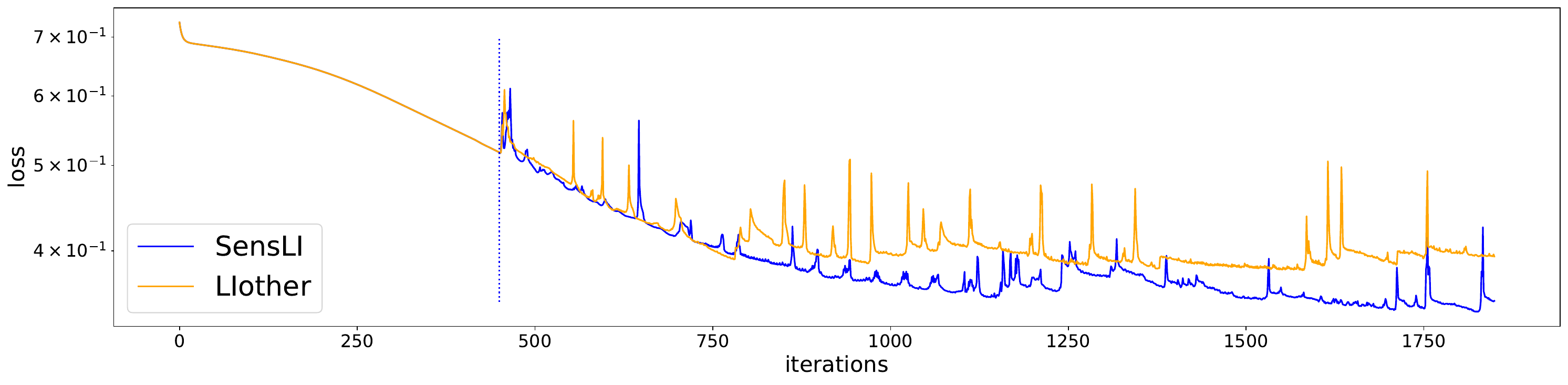}
		\caption{FNN, full-batch GD.}
		\label{figure:validity-of-layer-insertion-fnns}
	\end{subfigure}
	\hfill
	\begin{subfigure}[c]{\linewidth}
		\includegraphics[width = 0.98\linewidth, right]{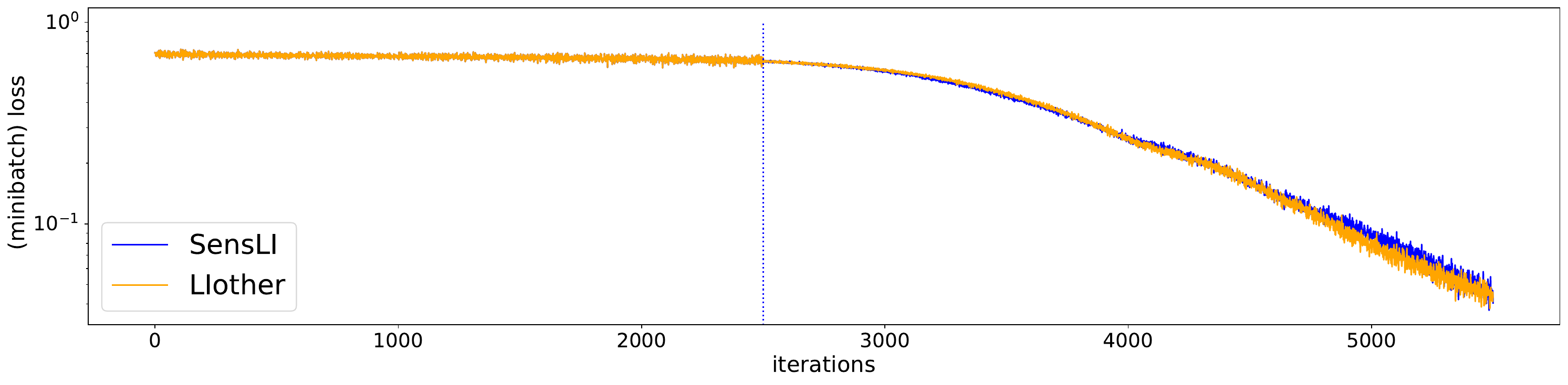}
		\caption{FNN, mini-batch SGD.}
		\label{figure:validity-of-layer-insertion-mb}
	\end{subfigure}
	\hfill
	\begin{subfigure}[c]{\linewidth}
		\includegraphics[width = \linewidth, right]{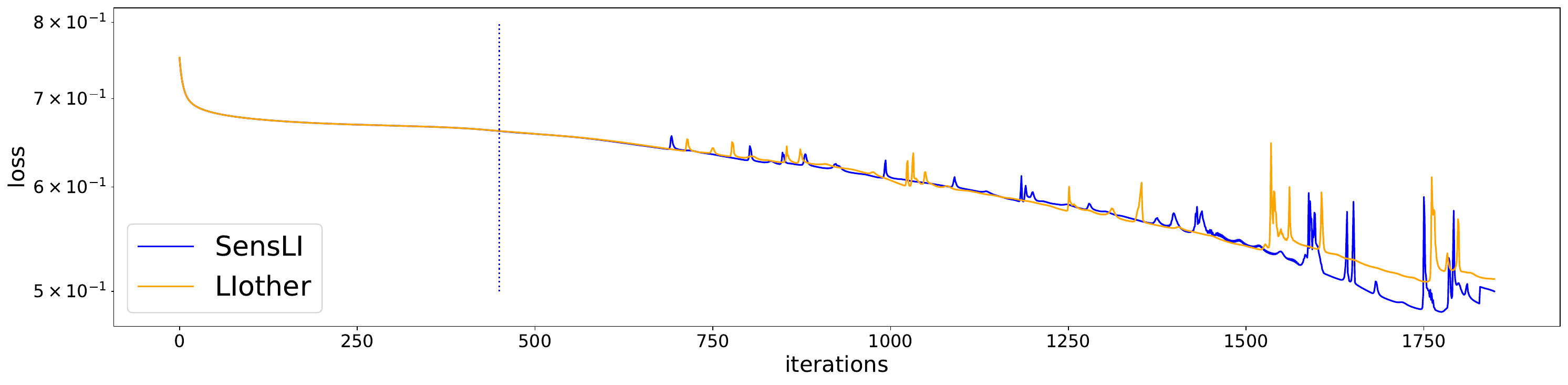}
		\caption{ResNet, full-batch GD.}
		\label{figure:validity-of-layer-insertion-resnets}
	\end{subfigure}
	\hfill
	\begin{subfigure}[c]{\linewidth}
		\includegraphics[width = 0.97\linewidth, right]{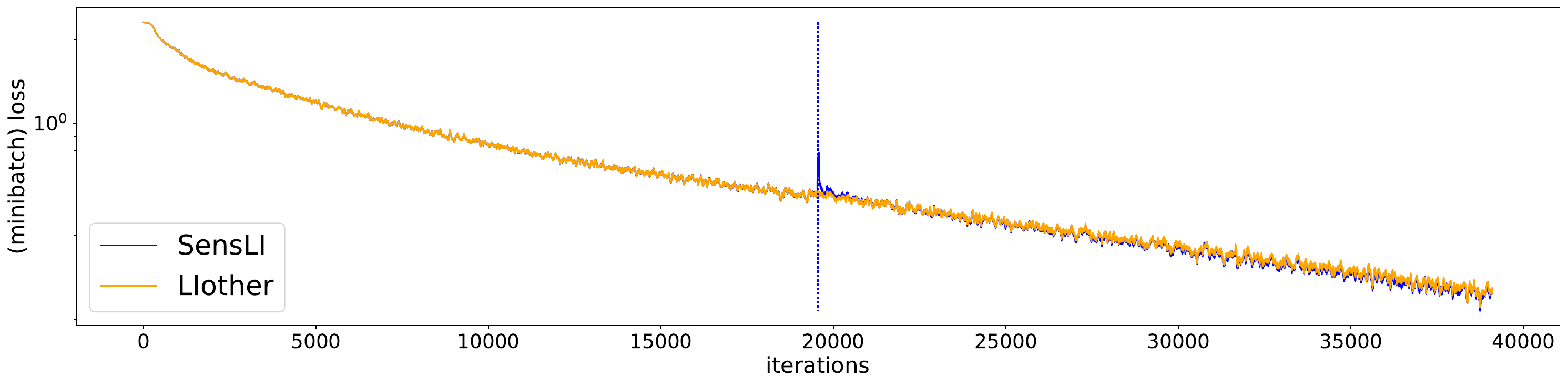}
		\caption{CNN, mini-batch SGD with momentum.}
		\label{figure:validity-of-layer-insertion-cnn}
	\end{subfigure}
	\caption{%
		Comparison of layer insertion at positions given by the largest (SensLI) and the smallest (LIother) of the merit indicators.
		The indicators for each layer are given by \eqref{eq:notion-of-merit-per-layer:ReLU-FNN} \wrt $W$ for FNNs (\subref{figure:validity-of-layer-insertion-fnns} and \subref{figure:validity-of-layer-insertion-mb}), \eqref{eq:notion-of-merit-per-layer:ReLU-FNN} \wrt $W_2$ for ResNets (\subref{figure:validity-of-layer-insertion-resnets}), and \eqref{eq:notion-of-merit-per-layer:CNN} for CNNs (\subref{figure:validity-of-layer-insertion-cnn}).
		We show the loss over iteration count, averaged over 30 (FNN, ResNet) and 7 (CNN) training runs.
		These experiments can be found in the \github repository as Exp6, Exp8, Exp7 and Exp9.
	}
	\label{figure:validity-of-layer-insertion}
\end{figure}

\begin{figure}[tb]
	\centering
	\begin{subfigure}[c]{\linewidth}
		\includegraphics[width = \linewidth, right]{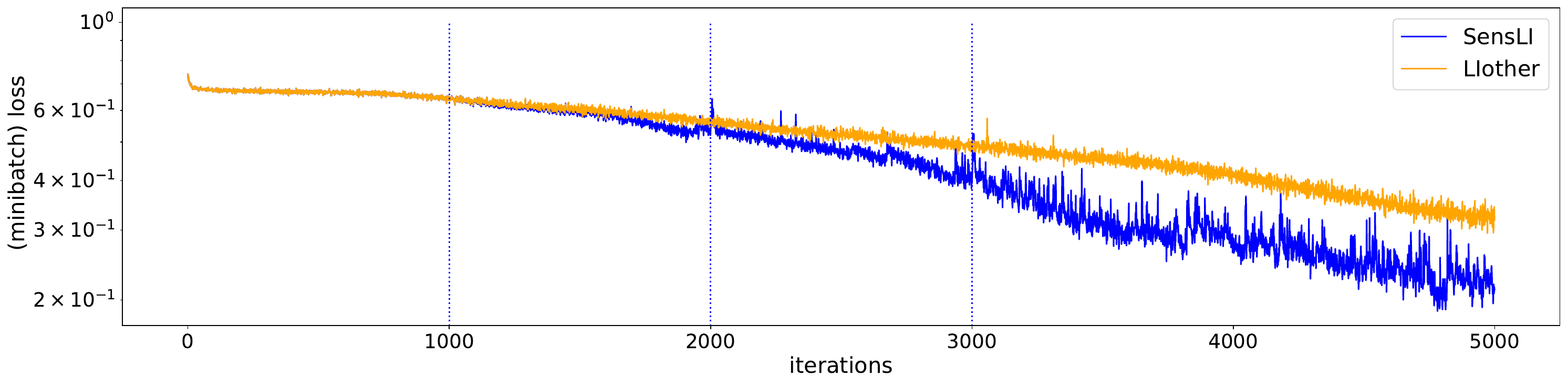}
		\caption{ResNet, mini-batch SGD.}
		\label{figure:validity-of-layer-insertion-resnets-3lis}
	\end{subfigure}
	\hfill
	\begin{subfigure}[c]{\linewidth}
		\includegraphics[width = 0.98\linewidth, right]{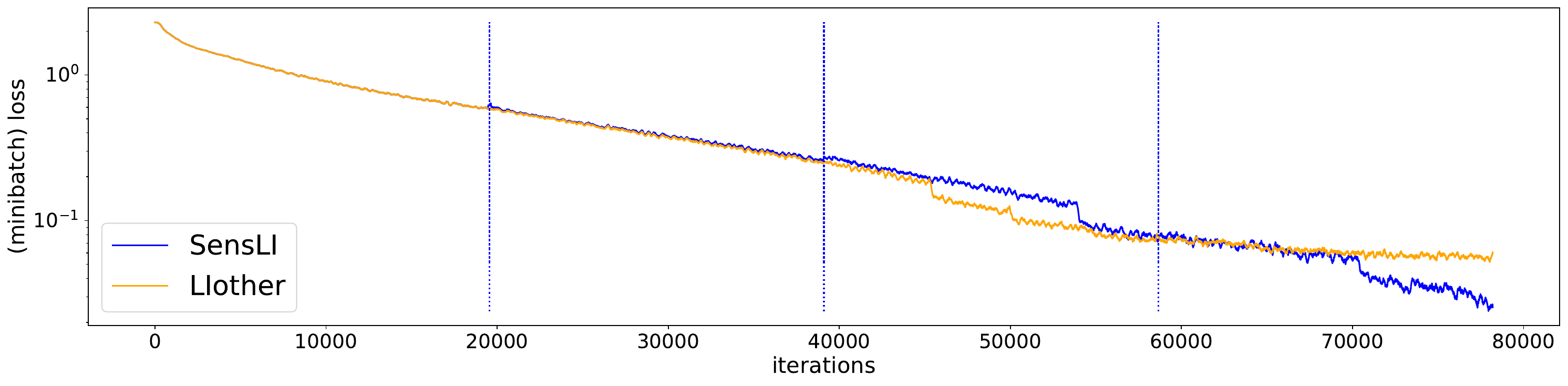}
		\caption{CNN, mini-batch SGD.}
		\label{figure:validity-of-layer-insertion-cnn-3lis}
	\end{subfigure}
	\caption{%
		Comparison of layer insertion at positions given by the largest (SensLI) and the smallest (LIother) of the merit indicators.
		A layer is inserted three times each in the training process.
		We show the loss over iteration count and test error over epochs, averaged over 30 (ResNet) and 7 (CNN) training runs.
		These experiments can be found in the \github repository as Exp13 and Exp14.
	}
	\label{figure:validity-of-layer-insertion-3lis}
\end{figure}

Although the proposed placement strategy is based entirely on local information, for full-batch training we observe a slight advantage of SensLI over LIother,
both close to insertion time and in the long run, \cf \cref{figure:validity-of-layer-insertion}.
We observe that the noise from mini-batch training can dominate the local effects of SensLI, as observed \eg in \cref{figure:validity-of-layer-insertion-cnn}.
However, for repeated layer insertions, the advantage of SensLI over LIother becomes visible again, \cf \cref{figure:validity-of-layer-insertion-3lis}.
We display two experiments with mini-batch SGD training.
Specifically, we consider a ResNet architecture in \cref{figure:validity-of-layer-insertion-resnets-3lis} and also compare LIother to our experimental setup from \cref{figure:performance-of-layer-insertion-cnn} in \cref{figure:validity-of-layer-insertion-cnn-3lis}.
Due to averaging over multiple runs, here, we plot over iterations instead of time.
We observe that the LIother strategy is less effective than SensLI to reduce the training loss despite  existing noise from the mini-batch training.

\FloatBarrier

\subsection{Comparison of SensLI to Other Layer Insertion Methods}
\label{subsection:comparison-to-other-layer-insertion-methods}

In this section we compare SensLI to other informed layer insertion methods and to random layer insertion.
Other methods which consider layer insertion and answer the question \enquote{Where to insert?} are SENN, Firefly and Autogrow, \cf \cref{tab:comparison}.
However, Autogrow employs random initialization and is not an informed choice.
Therefore, we only compare to SENN and Firefly here.

A fair numerical comparison is not straightforward for the following reasons:
SENN layer insertion for CNNs is only implemented for DenseNet architectures, while in the implementation of Firefly, layer insertion can not be executed separately from layer widening.
We therefore provide a theoretical comparison of the computational effort needed for the network expansion, which shows that SensLI is significantly less computationally expensive.

SensLI executes one full-batch forward and backward pass on the fully extended network (containing new layers in all possible positions).
A new layer is then selected based on the gradient norm \wrt the variables of the respective layer, \cf \cref{alg:sensli}.

SENN and Firefly are also both based on the idea of inserting a layer in all possible positions, but differ in the way they select the new layer.
Using SENN, $N$ random weight initializations are executed for $M$ iterations, leading to a computational effort of $N \times M$ (large) mini-batch forward and backward passes of a network extended by one layer.
Typical values are $N = 100$, $M = 300$ and a mini-batch size of $B=\num{1000}$ out of $T = \num{50000}$ training data points. This corresponds to $N \times M \times B/T = 600$ full-batch forward and backward passes of the partially extended network; see \cite[Appendix~B]{MitchellMenzenbachKerstingMundt:2023:1}.
For every random initialization a natural expansion score (approximated using a KFAC approximation of the Fisher matrix) is computed.
The position of the layer is chosen by comparing expansion scores of all possible positions, \ie, the number of forward and backward passes on the partially expanded networks has to be multiplied by the number of possible layer positions.

In Firefly \cite{WuLiuStoneLiu:2020:1}, initialization and new neuron positions are optimized simultaneously.
It performs $M$ full-batch gradient descent iterations on the fully extended network (containing new layers in all possible positions) with additional variables for each new neuron.
To execute these optimization steps $M$ full-batch forward and backward passes need to be computed.
The authors report that \enquote{a few} iterations suffice, \ie, $M \leq 10$.
New neurons are selected based on the values of the corresponding optimization variables.

Altogether, SensLI requires one full-batch forward and backward pass on the fully extended network, while SENN needs multiple hundred full-batch forward and backward passes on partially extended networks and for Firefly several optimization steps are executed that each contain a full-batch forward and backward pass.

\begin{figure}[tb]
	\centering
	\begin{subfigure}[c]{\linewidth}
		\includegraphics[width = \linewidth, right]{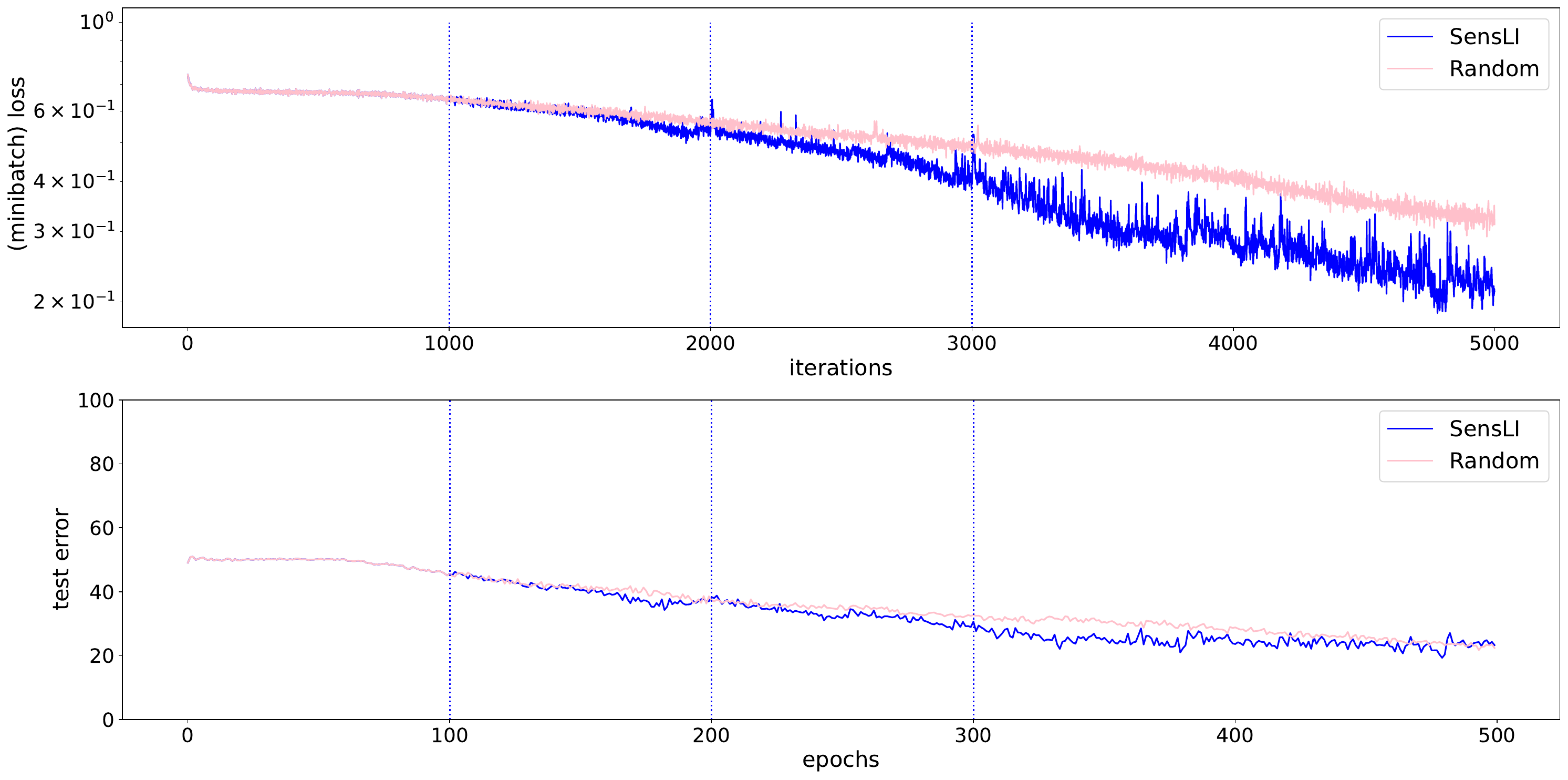}
		\caption{ResNet, mini-batch SGD.}
		\label{figure:vs-random-resnets-3lis}
	\end{subfigure}
	\hfill
	\begin{subfigure}[c]{\linewidth}
		\includegraphics[width = 0.98\linewidth, right]{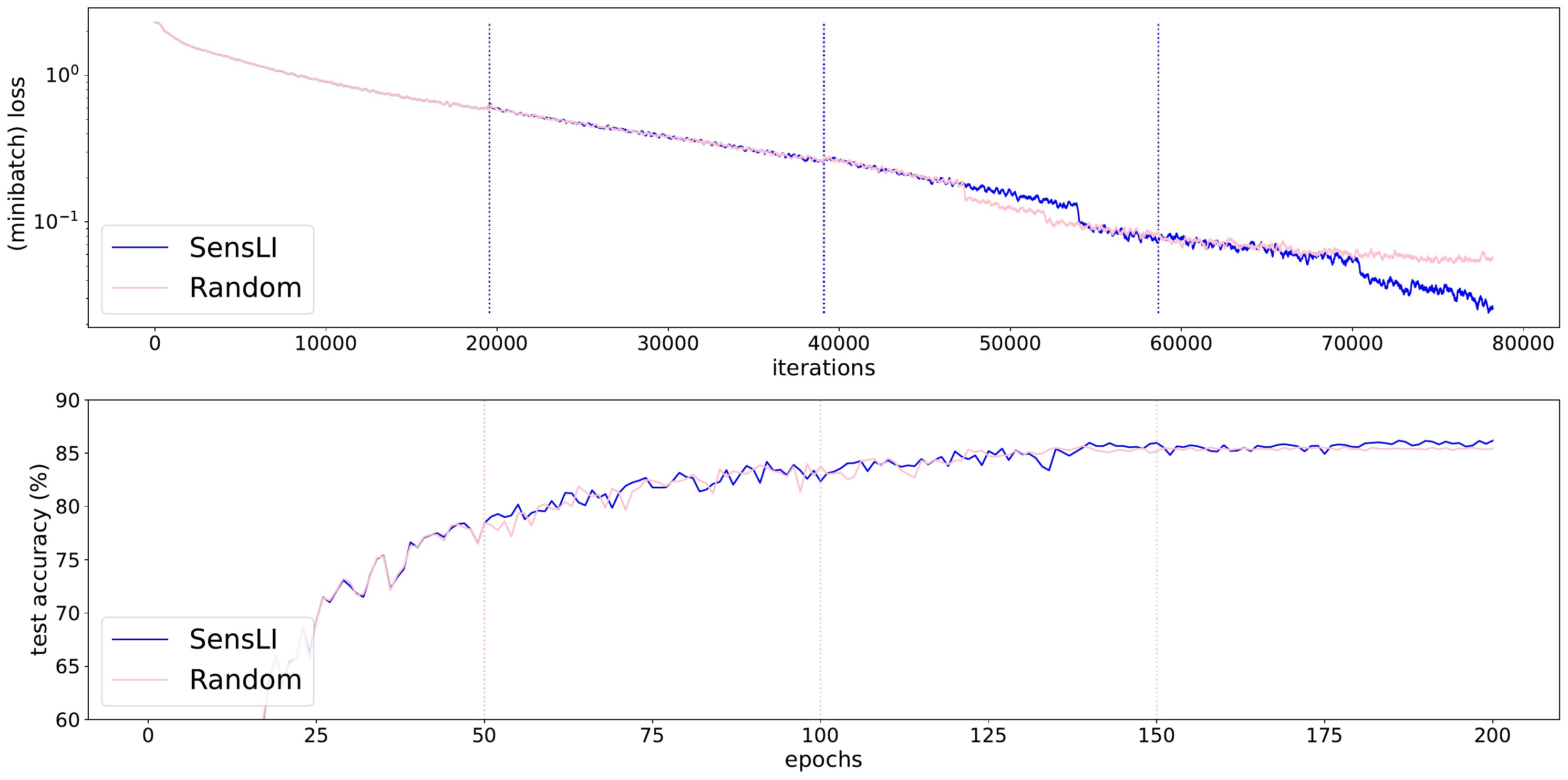}
		\caption{CNN, mini-batch SGD.}
		\label{figure:vs-random-cnn-3lis}
	\end{subfigure}
	\caption{%
		Comparison of layer insertion at positions given by the largest of the merit indicators (SensLI) or random positioning (Random).
		Layer insertion is executed three times in the training process, indicated by dotted vertical lines.
		We show the loss over iteration count and test error over epochs, averaged over 30 (ResNet) and 7 (CNN) training runs.
		These experiments can be found in the \github repository as Exp13 and Exp14.
	}
	\label{figure:vs-random-3lis}
\end{figure}

Since a fair numerical comparison to SENN and Firefly is not straightforward, we compare SensLI to a random layer insertion strategy instead, which randomly selects a layer position.
We show the results for ResNet and CNN architectures with three~subsequent layer insertions in \cref{figure:vs-random-3lis}.
Again we average over multiple training runs with different random initializations.
We see that SensLI outperforms the random insertion strategy in both cases.

\section{Conclusion}
\label{section:discussion}

An advantage of the SensLI approach to layer insertion during training is its simplicity.
When training is suspended for a potential layer insertion, we build a temporary, fully extended network by inserting layers in all possible locations, then evaluate the gradient of the extended objective once by backpropagation.
Hence, SensLI is computationally cheaper than currently used layer insertion methods such as SENN \citep{MitchellMenzenbachKerstingMundt:2023:1} and Firefly \citep{WuLiuStoneLiu:2020:1}, while still providing answers to the questions of where, how and when to insert a layer.
In contrast to other methods, SensLI is not relying on a costly selection of layers, or on uninformed weight initialization using a set of randomly generated instances.
SensLI provides a clear criterion for layer insertion rooted in sensitivity analysis for nonlinear programming and is able to insert residual as well as feedforward, fully connected and convolutional layers.
While, in our experiments, we focus on the insertion of a single layer at a time, the merit indicators \eqref{eq:notion-of-merit-per-layer:ReLU-FNN} and \eqref{eq:notion-of-merit-per-layer:CNN} can also be employed to insert multiple layers at once.

Although SensLI is based on a first-order and thus local prediction, we do observe global improvements in loss decay and test accuracy.
This is quite surprising since, using sensitivity analysis, it is impossible to make predictions with regard to the long-term effect of layer insertion, let alone predict which layer would lead to overall fastest convergence, or best accuracy on test data.

Preliminary theoretical results indicate that SensLI can be extended to a general network expansion strategy, including layer insertion and layer widening.
We leave the details to future research.

\FloatBarrier

\appendix
\section{Sensitivity Norm Computation}
\label{section:sensitivity-norms}

In CNNs, $\nabla_K \fextended(\thetaextended)$ is a higher-dimensional object than the gradient of a weight matrix in, \eg, a feedforward neural network.
It is therefore sensible to choose a norm suited to the structure of the kernel.
Different possibilities for options arise; choosing the Frobenius norm, \cf \eqref{eq:sensitivity-norms1}, as for the fully-connected networks is an uninformed choice.
Scaling the Frobenius norm by the number of elements in the kernel tensor, see \eqref{eq:sensitivity-norms2}, helps the comparison of layers with different kernel sizes.

Choosing an alternative viewpoint, we can consider the operator norm of the linear map $A_K$ representing the convolution operation with kernel $K$.
The indices $i$ and $j$ indicate the spatial filter positions, while $k$ and $\ell$ denote the input and output channels, respectively.
The operator norm is the maximum singular value of $A_K$, see \eqref{eq:notion-of-merit-per-layer:CNN}.
Using the method from \citet{SedghiGuptaLong:2018:1}, we can evaluate it with reasonable effort in $O(d^2 c^3)$ time,
where $d$ is the spatial size of the convolutional layers and $c$ is the number of input/output channels.
Measuring the impact of different output channels more specifically, we can consider the $2 \leftarrow 2$ norm over the input channels as a linear map
for each output channel separately and take the squared 2-norm, \eqref{eq:sensitivity-norms3}, or 1-norm, \eqref{eq:sensitivity-norms4}, over the output channels.
Altogether, we define
\begin{subequations}
	\begin{align}
		\label{eq:sensitivity-norms1}
		\norm{K}_F^2
		&
		\coloneqq
		\sum_{i,j,k,\ell} K^2_{i,j,k,\ell}
		,
		\\
		\label{eq:sensitivity-norms2}
		\norm{K}_{F,\textup{scaled}}^2
		&
		\coloneqq
		\frac{1}{c_i \, c_j \, c_k \, c_\ell} \sum_{i,j,k,\ell} K^2_{i,j,k,\ell}
		,
		\\
		\label{eq:notion-of-merit-per-layer:CNN}
		\norm{K}_{2 \leftarrow 2}^2
		&
		\coloneqq
		\sup_{\norm{x}_2 = 1} \norm[big]{A_K x}_2^2
		,
		\\
		\label{eq:sensitivity-norms3}
		\norm{K}_{2,2 \leftarrow 2}^2
		&
		\coloneqq
		\sum_{\ell\text{ outchannels}} \norm[Big]{\sum_{\mrep{k \text{ inchannels}}{\hspace*{11mm}}} K_{:,:,k,\ell}}_{2 \leftarrow 2}^2
		,
		\\
		\label{eq:sensitivity-norms4}
		\norm{K}_{1,2 \leftarrow 2}^2
		&
		\coloneqq
		\paren[Big](){\sum_{\ell\text{ outchannels}} \norm[Big]{\sum_{\mrep{k \text{ inchannels}}{\hspace*{11mm}}} K_{:,:,k,\ell}}_{2 \leftarrow 2}}^2
		,
	\end{align}
\end{subequations}
where $K$ is the kernel tensor, $c_i$ and $c_j$ are the spatial filter sizes, and $c_k$ and $c_\ell$ are the number of input and output channels, respectively.
Furthermore, $A_K$ is the matrix which represents the convolution operation with kernel $K$ as a linear map.

\begin{figure}[bt]
	\centering
	\includegraphics[width = 0.7\linewidth]{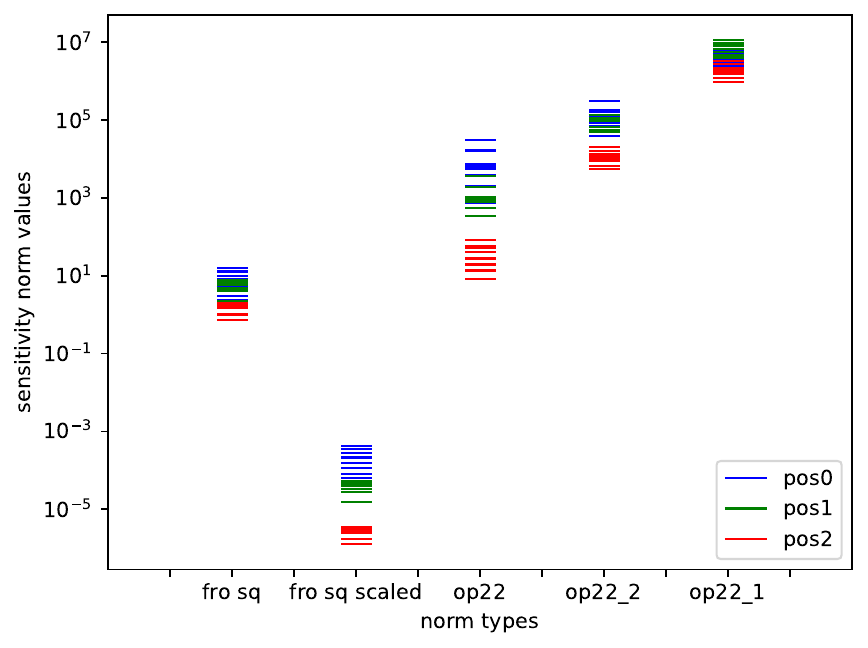}
	\caption{%
		Comparison of different norms for the sensitivity of layer insertion in CNNs.
		From left to right we show $\norm{K}_F^2$, \eqref{eq:sensitivity-norms1}, $\norm{K}_{F,\textup{scaled}}^2$, \eqref{eq:sensitivity-norms2}, $\norm{K}_{2 \leftarrow 2}^2$, \eqref{eq:notion-of-merit-per-layer:CNN}, $\norm{K}_{2,2 \leftarrow 2}^2$, \eqref{eq:sensitivity-norms3} and $\norm{K}_{1,2 \leftarrow 2}^2$, \eqref{eq:sensitivity-norms4}.
			We display 10 runs and every line per norm per position indicates the result from one run.
		The experiment can be found in the \github repository as Exp5.
	}
	\label{figure:comparison-of-sens-norms}
\end{figure}
We compare the behavior of the different norms numerically in \cref{figure:comparison-of-sens-norms} for $10$ runs of the same experiment.
Details of the experiment can be found in \cref{section:hyperparameters}.
We consider a CNN with one additional layer inserted after 50 epochs and observe, that in this setup, the training process behaves very similarly for each individual run and the same position, specifically position 0, is favorable for layer insertion.
The five norms have different orders of magnitude compared to each other and some choices do not provide a clear separation between the different positions of the layer insertion, \cf \cref{figure:comparison-of-sens-norms}.
The operator norm is the most suitable for our strategy, as it provides a clear separation between the different positions of the layer insertion.
The Frobenius norm scaled by the number of elements in the kernel tensor is a reasonable alternative.

\section{Comparison of Layer Insertion Points}
\label{section:numerical-when-to-insert}

In this experiment, we investigate the impact of inserting a layer at different times during the training progress.
In all experiments we confine ourselves to adding a single layer.
Since we restore the random seed, the initialization for all runs where we insert a layer during training is the same as for the baseline network (FNN1/CNN1).
Consequently, when we use full-batch gradient descent, the training histories all coincide with that of FNN1 until the new layer is inserted.
Considering the training algorithms full-batch SGD, mini-batch SGD, and mini-batch-SGD with momentum once each, we compare the training histories of the fixed-architecture baseline network (FNN1/CNN1) and the networks obtained from inserting the additional layer at various points during training.
We examine eight options, respectively, \ie, insertion at iteration $150, 250, \ldots, 850$ for the FNN full-batch SGD training, insertion at epoch $50,100,\ldots,400$ for the FNN mini-batch SGD training
and insertion at epoch $10,20,\ldots,80$ for the CNN mini-batch SGD with momentum training.
To ensure comparability, we use the same hyperparameters for all layer insertions.
The experimental details can be found in \cref{section:hyperparameters}.

The training histories are shown in \cref{figure:when-to-insert-layers}.
From this experiment it becomes apparent that finding the best time when to insert a new layer is not obvious.
Evidently, in this experiment, the insertion of a layer during training at any point out of the eight options compares favorably to not adding a second hidden layer (FNN1/CNN1).
However, we see that inserting a layer too late renders it less effective.

For the FNN full-batch example and random instance at hand, we find that the most effective point out of the eight options tested is after 450~iterations.
For the FNN mini-batch example, the training histories are more monotone and the earlier the layer is inserted, the faster the loss decreases.
For the CNN mini-batch with momentum example, the training histories are more erratic and the choice of the insertion point has less impact.
We further observe that the later the layer is inserted in the training process, the more it disrupts the training process.
We remark that at each insertion epoch, the method chooses the same position for the layer insertion for the CNN.

\begin{figure}[htb]
	\centering
	\begin{subfigure}{\columnwidth}
		\centering
		\includegraphics[width = 0.85\columnwidth]{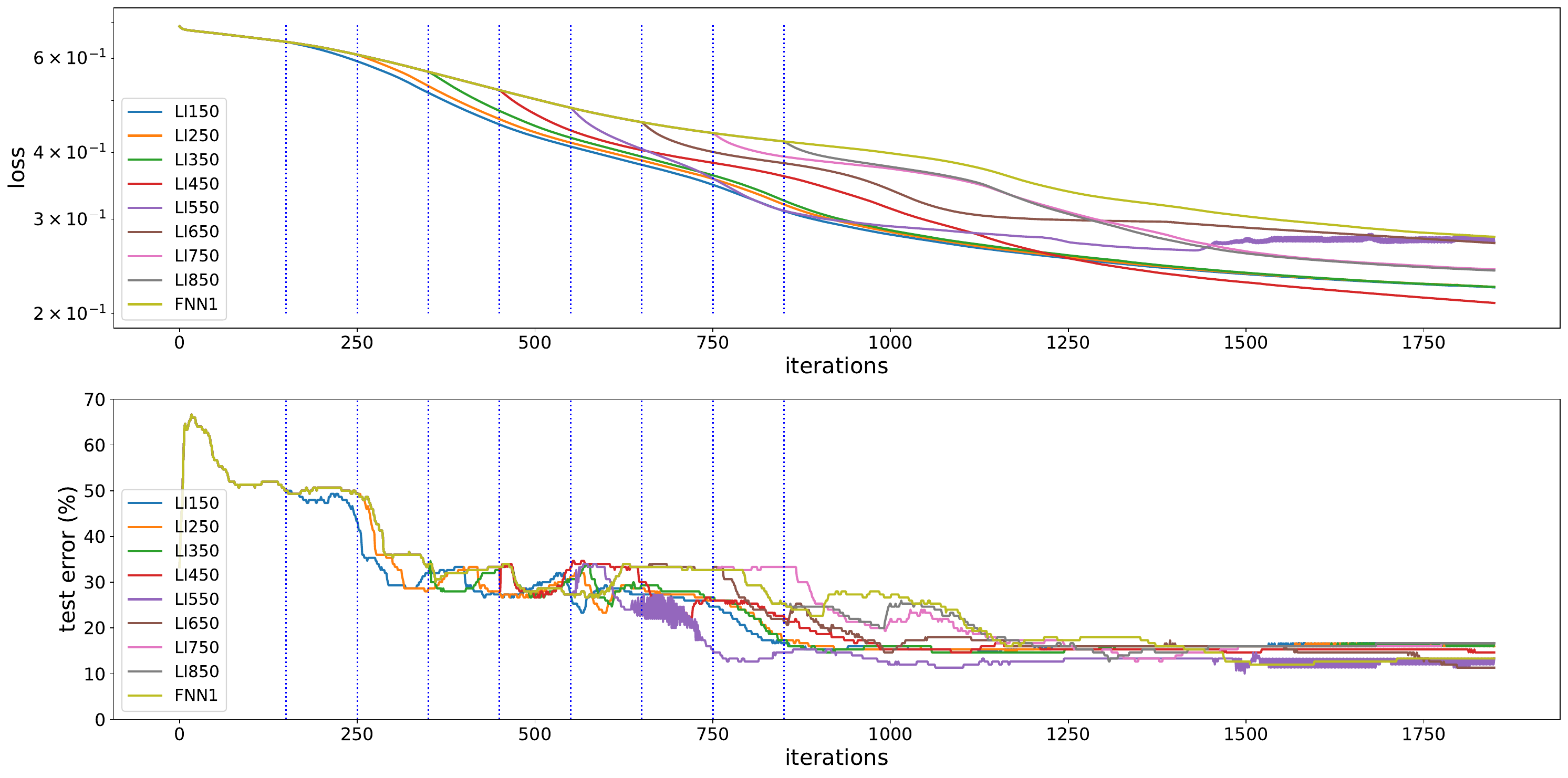}
		\caption{FNN, full-batch GD.}
		\label{figure:when-to-insert-layers-fnn-fb}
	\end{subfigure}
\end{figure}%

\begin{figure}[ht]\ContinuedFloat
	\begin{subfigure}{\columnwidth}
		\centering
		\includegraphics[width = 0.85\columnwidth]{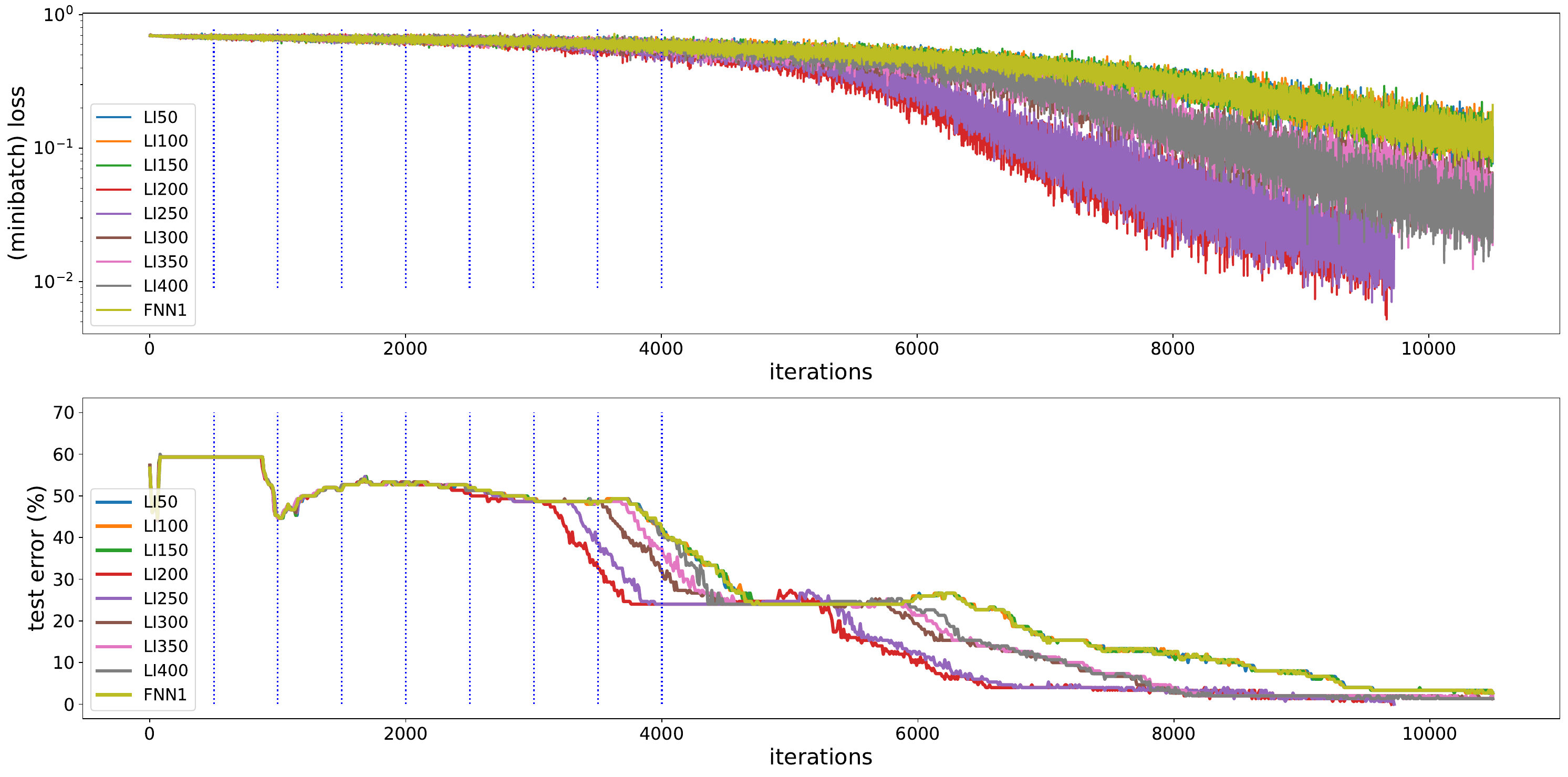}
		\caption{FNN, mini-batch SGD.}
		\label{figure:when-to-insert-layers-fnn-mb}
	\end{subfigure}
\end{figure}%

\begin{figure}[ht]\ContinuedFloat
	\begin{subfigure}{\columnwidth}
		\centering
		\includegraphics[width = 0.85\columnwidth]{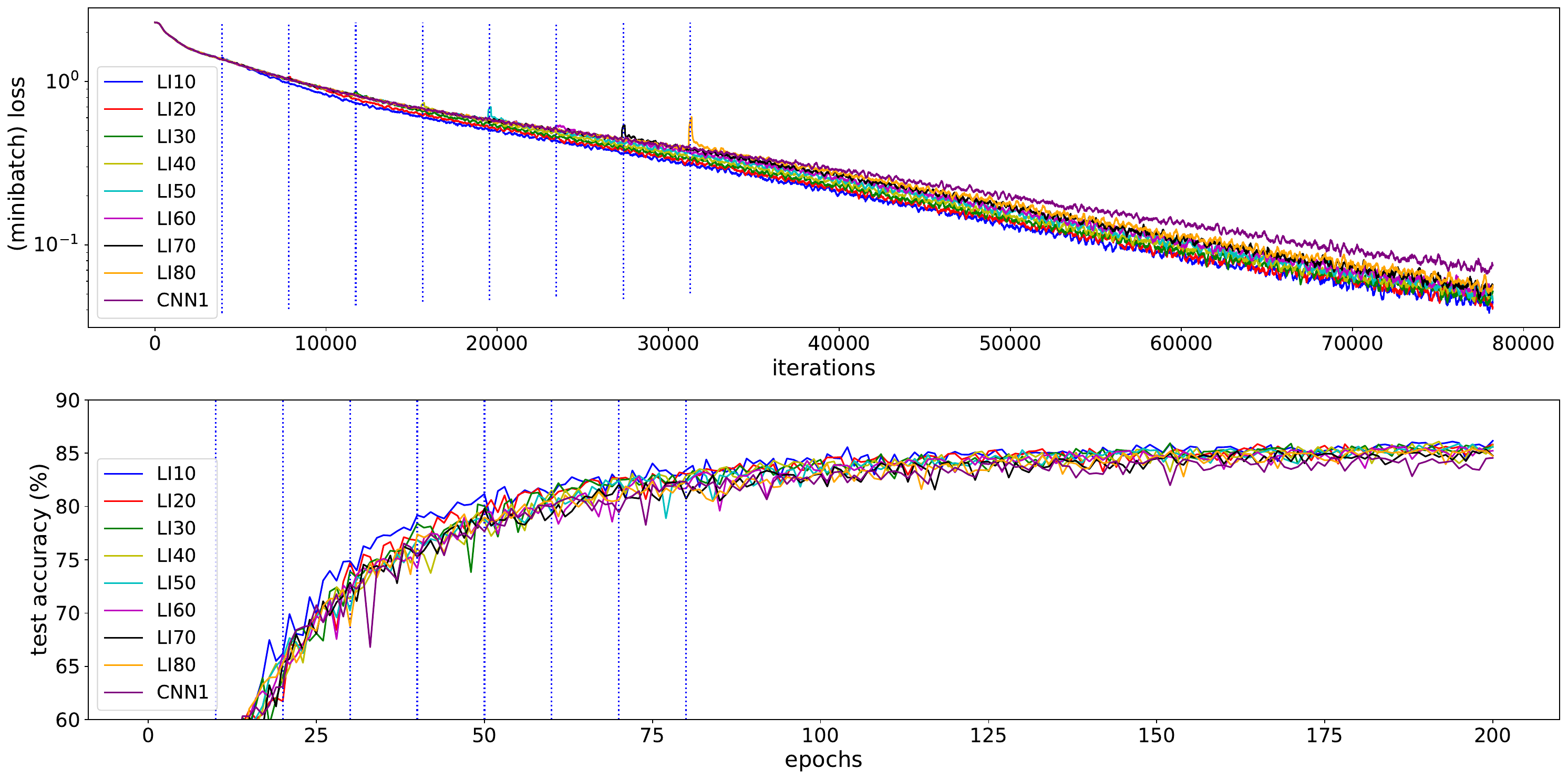}
		\caption{CNN, mini-batch SGD with momentum.}
		\label{figure:when-to-insert-layers-cnn}
	\end{subfigure}
	\caption{%
		Comparison of layer insertion at different iterations (indicated by vertical lines) as described in \cref{section:numerical-when-to-insert}.
		We show the loss and test error over iteration count.
		These experiments can be found in the \github repository under the name Exp10 and Exp11 for the \relu-FNN and Exp12 for the CNN.
	}
	\label{figure:when-to-insert-layers}
\end{figure}

\FloatBarrier

\section{Detailed Experiment Setup}
\label{section:hyperparameters}

Generally, we use the standard initialization from \pytorch for the parameters of the networks. Below, we list the detailed setup, including hyperparameter specifications for all experiments. Additionally, it is indicated, which figures display results from the respective experiment.

\paragraph{Experiment 1 (CNN)}

\begin{itemize}
	\item \textbf{Figures:} \cref{figure:performance-of-layer-insertion-cnn}
	\item \textbf{Data set:} CIFAR-10.
	\item \textbf{Data set size} \num{60000}.
	\item \textbf{Train/Test split} \num{50000}/\num{10000}.
	\item \textbf{Data augmentation} RandomCrop, RandomHorizontalFlip, Normalize.
	\item \textbf{Batch size:} 128.
	\item \textbf{Architecture (CNN LI):} Baseline network as described in \cref{figure:CNN-architecture}, then extended by SensLI. Total number of parameters: as in \cref{table:parameters-cnn}.
	\item \textbf{Architecture (CNN):} Baseline network extended with two additional layers with channel numbers 64x64 and 64x64. Total number of parameters: as in \cref{table:parameters-cnn}.
	\item \textbf{Activation function:} \relu.
	\item \textbf{Loss function:} Cross-entropy loss.
	\item \textbf{Optimizer:} Adam with learning rate $0.01$.
	\item \textbf{Learning rate schedule:} ReduceLROnPlateau with patience $10$ and factor $0.5$.
	\item \textbf{Number of layer insertions by SensLI:} 3.
	\item \textbf{Training epochs total:} 200.
	\item \textbf{Layer insertion:} After every 50 epochs.
	\item \textbf{Layer insertion threshold:} as described in \cref{subsection:when-to-insert} with $\tau=1$.
	\item \textbf{Number of runs:} 1.
	\item \textbf{Decrease of learning rate after layer insertion:} None.
	\item \textbf{Smoothing in Figures:} Moving average with window size 170.
\end{itemize}
\begin{table}[h]
	\caption{Number of network parameters during training for the respective networks compared in \cref{figure:performance-of-layer-insertion-cnn}.}
	\label{table:parameters-cnn}
	\centering
		\begin{tabular}{rrrrrrr}
			\toprule
			Epochs  & 0--49         & 50--99        & 100--149      & 150--199
			\\
			\midrule
			SensLI  & \num{2674816} & \num{2711744} & \num{2711744} & \num{2748672}
			\\
			CNN     & \multicolumn{4}{c}{\num{2748672}}
			\\
			\bottomrule
		\end{tabular}
\end{table}

\paragraph{Experiment 2 (FNN)}

\begin{itemize}
	\item \textbf{Figures:} \cref{figure:comparison-with-fixed-architecture-fnns}
	\item \textbf{Data set:} Spirals see \cref{figure:illustration-spiral-dataset}
	\item \textbf{Data set size} 600.
	\item \textbf{Train/Test split} 450/150.
	\item \textbf{Data augmentation} None.
	\item \textbf{Batch size:} 450.
	\item \textbf{Architecture (FNN LI):} Fully-connected feedforward network with one hidden layer of width 5, then extended by SensLI. Total number of parameters: as in \cref{table:parameters-fb}.
	\item \textbf{Architecture (FNN1):} Fully-connected feedforward network with one hidden layer of width 5.  Total number of parameters: 27.
	\item \textbf{Architecture (FNN2):} Fully-connected feedforward network with two hidden layers of width 5. Total number of parameters: 57.
	\item \textbf{Activation function:} \relu.
	\item \textbf{Loss function:} Cross-entropy loss.
	\item \textbf{Optimizer:} full-batch GD with learning rate $0.01$.
	\item \textbf{Learning rate schedule:} fixed learning rate.
	\item \textbf{Number of layer insertions by SensLI:} 1.
	\item \textbf{Training epochs total:} \num{1850}.
	\item \textbf{Layer insertion:} After 450 epochs.
	\item \textbf{Number of runs:} 30.
	\item \textbf{Decrease of learning rate after layer insertion:} None.
	\item \textbf{Smoothing in Figures:} None.
\end{itemize}

\paragraph{Experiment 3 (ResNet)}

\begin{itemize}
	\item \textbf{Figures:} \cref{figure:comparison-with-fixed-architecture-resnets}
	\item \textbf{Data set:} Spirals see \cref{figure:illustration-spiral-dataset}
	\item \textbf{Data set size} 600.
	\item \textbf{Train/Test split} 450/150.
	\item \textbf{Data augmentation} None.
	\item \textbf{Batch size:} 450.
	\item \textbf{Architecture (ResNet LI):} ResNet as defined in \cref{eq:ResNet} with 2 hidden layers of width 3, then extended by SensLI. Total number of parameters: as in \cref{table:parameters-fb}.
	\item \textbf{Architecture (ResNet1):} ResNet as defined in \cref{eq:ResNet} with 2 hidden layers of width 3. Total number of parameters: 33.
	\item \textbf{Architecture (ResNet2):} ResNet as defined in \cref{eq:ResNet} with 3 hidden layers of width 3. Total number of parameters: 54.
	\item \textbf{Activation function:} $\tanh$.
	\item \textbf{Loss function:} Cross-entropy loss.
	\item \textbf{Optimizer:} full-batch GD with learning rate $0.1$.
	\item \textbf{Learning rate schedule:} fixed learning rate.
	\item \textbf{Number of layer insertions by SensLI:} 1.
	\item \textbf{Training epochs total:} \num{1850}.
	\item \textbf{Layer insertion:} After 450 epochs.
	\item \textbf{Initialization of new layer:} $W_1 \coloneqq 0.8 \, \id_{h_k \times h_k}$ and $b \coloneqq 0$.
	\item \textbf{Number of runs:} 30.
	\item \textbf{Decrease of learning rate after layer insertion:} None.
	\item \textbf{Smoothing in Figures:} None.
\end{itemize}

\paragraph{Experiment 4 (FNN)}

\begin{itemize}
	\item \textbf{Figures:} \cref{figure:comparison-with-fixed-architecture-mb}, \cref{figure:LI-layerwise-grads-mb}
	\item \textbf{Data set:} Spirals see \cref{figure:illustration-spiral-dataset}
	\item \textbf{Data set size} 600.
	\item \textbf{Train/Test split} 450/150.
	\item \textbf{Data augmentation} None.
	\item \textbf{Batch size:} 45.
	\item \textbf{Architecture (FNN LI):} Fully-connected feedforward network with one hidden layer of width 5, then extended by SensLI. Total number of parameters: as in \cref{table:parameters-fb}.
	\item \textbf{Architecture (FNN1):} Fully-connected feedforward network with one hidden layer of width 5. Total number of parameters: 27.
	\item \textbf{Architecture (FNN2):} Fully-connected feedforward network with two hidden layers of width 5. Total number of parameters: 57.
	\item \textbf{Activation function:} \relu.
	\item \textbf{Loss function:} Cross-entropy loss.
	\item \textbf{Optimizer:} SGD with learning rate $0.01$.
	\item \textbf{Learning rate schedule:} fixed learning rate.
	\item \textbf{Number of layer insertions by SensLI:} 1.
	\item \textbf{Training epochs total:} \num{1850}.
	\item \textbf{Layer insertion:} After 450 epochs.
	\item \textbf{Number of runs:} 40.
	\item \textbf{Decrease of learning rate after layer insertion:} None.
	\item \textbf{Smoothing in Figures:} None.
\end{itemize}

\paragraph{Experiment 5 (CNN)}

\begin{itemize}
	\item \textbf{Figures:} \cref{figure:comparison-with-fixed-architecture-cnn}, \cref{figure:LI-layerwise-grads-cnn}, \cref{figure:comparison-of-sens-norms}
	\item \textbf{Data set:} CIFAR-10.
	\item \textbf{Data set size} \num{60000}.
	\item \textbf{Train/Test split} \num{50000}/\num{10000}.
	\item \textbf{Data augmentation} RandomCrop, RandomHorizontalFlip, Normalize.
	\item \textbf{Batch size:} 128.
	\item \textbf{Architecture (CNN LI):} Baseline network as described in \cref{figure:CNN-architecture}, then extended by SensLI. Total number of parameters: \num{2674816}, then \num{2711744}
	\item \textbf{Architecture (CNN1):} Baseline network as described in \cref{figure:CNN-architecture}. Total number of parameters: \num{2674816}
	\item \textbf{Architecture (CNN2):} Baseline network extended with one additional layers with channel numbers 64x64. Total number of parameters: \num{2711744}
	\item \textbf{Activation function:} \relu.
	\item \textbf{Loss function:} Cross-entropy loss.
	\item \textbf{Optimizer:} SGD with momentum $0.9$, weight decay $0.005$ and learning rate $0.01$.
	\item \textbf{Learning rate schedule:} fixed learning rate.
	\item \textbf{Number of layer insertions by SensLI:} 1.
	\item \textbf{Training epochs total:} 100.
	\item \textbf{Layer insertion:} After 50 epochs.
	\item \textbf{Number of runs:} 10.
	\item \textbf{Decrease of learning rate after layer insertion:} None.
	\item \textbf{Smoothing in Figures:} Moving average with window size 170 for \cref{figure:comparison-with-fixed-architecture-cnn}
		and 100 for \cref{figure:LI-layerwise-grads-cnn}.
\end{itemize}

\paragraph{Experiment 6 (FNN)}

\begin{itemize}
	\item \textbf{Figures:} \cref{figure:validity-of-layer-insertion-fnns}
	\item \textbf{Data set:} Spirals see \cref{figure:illustration-spiral-dataset}
	\item \textbf{Data set size} 600.
	\item \textbf{Train/Test split} 450/150.
	\item \textbf{Data augmentation} None.
	\item \textbf{Batch size:} 450.
	\item \textbf{Architecture (SensLI):} Fully-connected feedforward network with 2 hidden layers of width 4, then extended by SensLI. Total number of parameters: 42 before layer insertion, 62 after layer insertion.
	\item \textbf{Architecture (LIother):} Fully-connected feedforward network with 2 hidden layers of width 4, then extended by layer with smallest sensitivity. Total number of parameters: 42 before layer insertion, 62 after layer insertion.
	\item \textbf{Activation function:} \relu.
	\item \textbf{Loss function:} Cross-entropy loss.
	\item \textbf{Optimizer:} full-batch GD with learning rate $0.01$.
	\item \textbf{Learning rate schedule:} fixed learning rate.
	\item \textbf{Number of layer insertions by SensLI:} 1.
	\item \textbf{Training epochs total:} \num{1850}.
	\item \textbf{Layer insertion:} After 450 epochs.
	\item \textbf{Number of runs:} 30.
	\item \textbf{Decrease of learning rate after layer insertion:} None.
	\item \textbf{Smoothing in Figures:} None.
\end{itemize}

\begin{figure}[ht]
	\centering
	\includegraphics[width = 0.5\columnwidth]{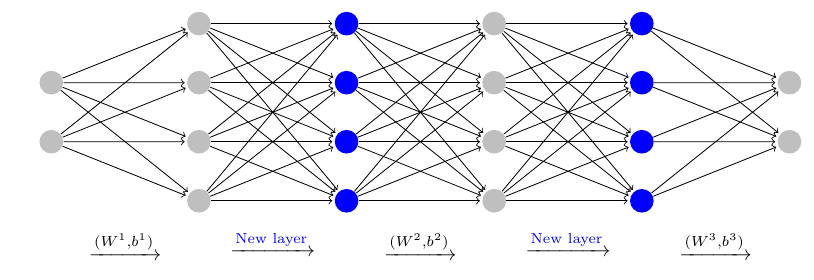}
	\caption{Exp6: The possible layer positions of SensLI for the architecture are shown in blue.}
	\label{figure:possible-positions-exp6}
\end{figure}

\paragraph{Experiment 7 (ResNet)}

\begin{itemize}
	\item \textbf{Figures:} \cref{figure:validity-of-layer-insertion-resnets}
	\item \textbf{Data set:} Spirals see \cref{figure:illustration-spiral-dataset}
	\item \textbf{Data set size} 600.
	\item \textbf{Train/Test split} 450/150.
	\item \textbf{Data augmentation} None.
	\item \textbf{Batch size:} 450.
	\item \textbf{Architecture (SensLI):} ResNet as defined in \cref{eq:ResNet} with 3 hidden layers of width 3, then extended by SensLI. Total number of parameters: 54 before layer insertion, 75 after layer insertion.
	\item \textbf{Architecture (LIother):} ResNet as defined in \cref{eq:ResNet} with 3 hidden layers of width 3, then extended by layer with smallest sensitivity. Total number of parameters: 54 before layer insertion, 75 after layer insertion.
	\item \textbf{Activation function:} $\tanh$.
	\item \textbf{Loss function:} Cross-entropy loss.
	\item \textbf{Optimizer:} full-batch GD with learning rate $0.1$.
	\item \textbf{Learning rate schedule:} fixed learning rate.
	\item \textbf{Number of layer insertions by SensLI:} 1.
	\item \textbf{Training epochs total:} \num{1850}.
	\item \textbf{Layer insertion:} After 450 epochs.
	\item \textbf{Initialization of new layer:} $W_1 \coloneqq 0.8 \, \id_{h_k \times h_k}$ and $b \coloneqq 0$.
	\item \textbf{Number of runs:} 30.
	\item \textbf{Decrease of learning rate after layer insertion:} factor $0.7$.
	\item \textbf{Smoothing in Figures:} None.
\end{itemize}

\begin{figure}[ht]
	\centering
	\includegraphics[width = 0.5\columnwidth]{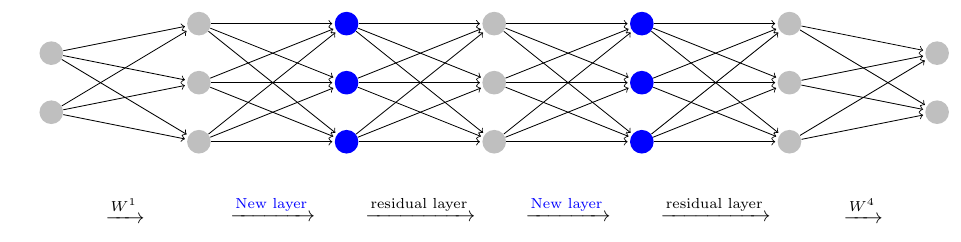}
	\caption{Exp7: The possible layer positions of SensLI for the architecture are shown in blue.}
	\label{figure:possible-positions-exp7}
\end{figure}

\paragraph{Experiment 8 (FNN)}

\begin{itemize}
	\item \textbf{Figures:} \cref{figure:validity-of-layer-insertion-mb}
	\item \textbf{Data set:} Spirals see \cref{figure:illustration-spiral-dataset}
	\item \textbf{Data set size} 600.
	\item \textbf{Train/Test split} 450/150.
	\item \textbf{Data augmentation} None.
	\item \textbf{Batch size:} 45.
	\item \textbf{Architecture (SensLI):} Fully-connected feedforward network with 2 hidden layers of width 10, then extended by SensLI. Total number of parameters: 162 before layer insertion, 272 after layer insertion.
	\item \textbf{Architecture (LIother):} Fully-connected feedforward network with 2 hidden layers of width 10, then extended by layer with smallest sensitivity. Total number of parameters: 162 before layer insertion, 272 after layer insertion.
	\item \textbf{Activation function:} \relu.
	\item \textbf{Loss function:} Cross-entropy loss.
	\item \textbf{Optimizer:} SGD with learning rate $0.01$ for epoch 0-250 and $0.02$ for epoch 250-550 for both FNNLI and LIother.
	\item \textbf{Learning rate schedule:} fixed learning rate.
	\item \textbf{Number of layer insertions by SensLI:} 1.
	\item \textbf{Training epochs total:} 550.
	\item \textbf{Layer insertion:} After 250 epochs.
	\item \textbf{Number of runs:} 30.
	\item \textbf{Decrease of learning rate after layer insertion:} As described in optimizer.
	\item \textbf{Smoothing in Figures:} None.
\end{itemize}

\begin{figure}[ht]
	\centering
	\includegraphics[width = 0.5\columnwidth]{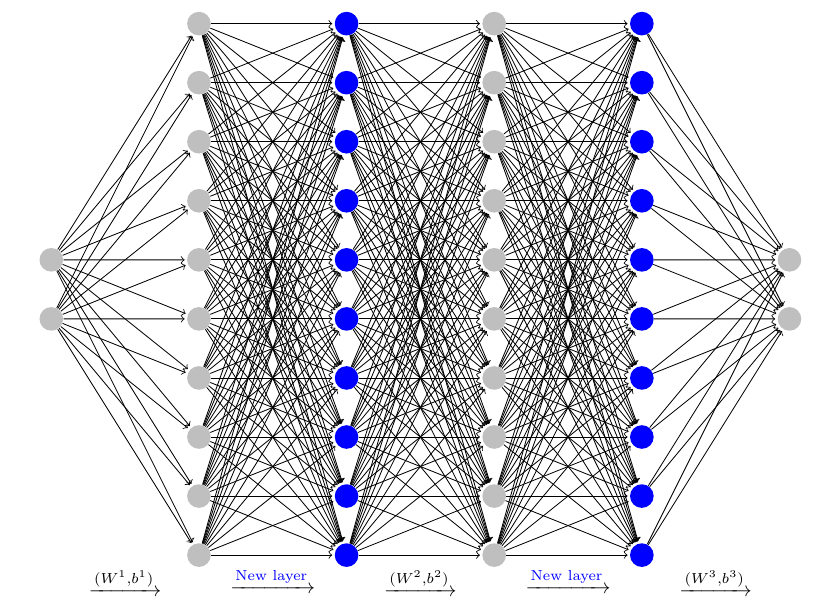}
	\caption{Exp8: The possible layer positions of SensLI for the architecture are shown in blue.}
	\label{figure:possible-positions-exp8}
\end{figure}

\FloatBarrier
\paragraph{Experiment 9 (CNN)}

\begin{itemize}
	\item \textbf{Figures:} \cref{figure:validity-of-layer-insertion-cnn}
	\item \textbf{Data set:} CIFAR-10.
	\item \textbf{Data set size} \num{60000}.
	\item \textbf{Train/Test split} \num{50000}/\num{10000}.
	\item \textbf{Data augmentation} RandomCrop, RandomHorizontalFlip, Normalize.
	\item \textbf{Batch size:} 128.
	\item \textbf{Architecture (SensLI):} Baseline network where each convolutional layer has $200$ channels (as opposed to \num{64128} and 256 in Exp30), then extended by SensLI. Total number of parameters: \num{2582000}, then \num{2942200}.
	\item \textbf{Architecture (LIother):} Baseline network where each convolutional layer has $200$ channels (as opposed to \num{64128} and 256 in Exp30), then extended by layer with smallest sensitivity. Total number of parameters: \num{2582000}, then \num{2942200}.
	\item \textbf{Loss function:} Cross-entropy loss.
	\item \textbf{Optimizer:} SGD with momentum $0.9$, weight decay $0.005$ and learning rate $0.01$.
	\item \textbf{Learning rate schedule:} fixed learning rate.
	\item \textbf{Number of layer insertions by SensLI:} 1.
	\item \textbf{Training epochs total:} 100.
	\item \textbf{Layer insertion:} After 50 epochs.
	\item \textbf{Number of runs:} 10.
	\item \textbf{Decrease of learning rate after layer insertion:} None.
	\item \textbf{Smoothing in Figures:} Moving average with window size 50.
\end{itemize}

\begin{figure}[ht]
	\centering
	\includegraphics[width = \columnwidth]{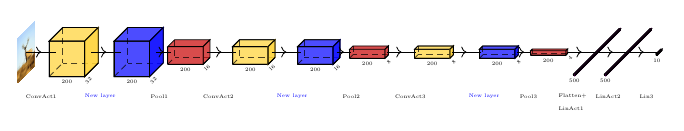}
	\caption{Exp9: The possible layer positions of SensLI for the architecture are shown in blue.}
	\label{figure:possible-positions-exp9}
\end{figure}

\FloatBarrier
\paragraph{Experiment 10 (FNN)}

\begin{itemize}
	\item \textbf{Figures:} \cref{figure:when-to-insert-layers-fnn-fb}
	\item \textbf{Data set:} Spirals see \cref{figure:illustration-spiral-dataset}
	\item \textbf{Data set size} 600.
	\item \textbf{Train/Test split} 450/150.
	\item \textbf{Data augmentation} None.
	\item \textbf{Batch size:} 450.
	\item \textbf{Architecture (FNN LI):} Fully-connected feedforward network with one hidden layer of width 5, then extended by SensLI. Total number of parameters: as in \cref{table:parameters-fb}.
	\item \textbf{Architecture (FNN1):} Fully-connected feedforward network with one hidden layer of width 5. Total number of parameters: 27.
	\item \textbf{Activation function:} \relu.
	\item \textbf{Loss function:} Cross-entropy loss.
	\item \textbf{Optimizer:} full-batch GD with learning rate $0.2$.
	\item \textbf{Learning rate schedule:} fixed learning rate.
	\item \textbf{Number of layer insertions by SensLI:} 1.
	\item \textbf{Training epochs total:} \num{1850}.
	\item \textbf{Layer insertion:} After $150, 250, \ldots, 850$ epochs.
	\item \textbf{Number of runs:} 1.
	\item \textbf{Decrease of learning rate after layer insertion:} factor $0.8$.
	\item \textbf{Smoothing in Figures:} None.
\end{itemize}

\paragraph{Experiment 11 (FNN)}

\begin{itemize}
	\item \textbf{Figures:} \cref{figure:when-to-insert-layers-fnn-mb}
	\item \textbf{Data set:} Spirals see \cref{figure:illustration-spiral-dataset}
	\item \textbf{Data set size} 600.
	\item \textbf{Train/Test split} 450/150.
	\item \textbf{Data augmentation} None.
	\item \textbf{Batch size:} 45.
	\item \textbf{Architecture (FNN LI):} Fully-connected feedforward network with 2 hidden layers of width 10, then extended by SensLI. Total number of parameters: 162 before layer insertion, 272 after layer insertion.
	\item \textbf{Architecture (FNN1):} Fully-connected feedforward network with 2 hidden layers of width 10.  Total number of parameters: 162.
	\item \textbf{Activation function:} \relu.
	\item \textbf{Loss function:} Cross-entropy loss.
	\item \textbf{Optimizer:} SGD with learning rate $0.01$.
	\item \textbf{Learning rate schedule:} fixed learning rate.
	\item \textbf{Number of layer insertions by SensLI:} 1.
	\item \textbf{Training epochs total:} \num{1050}.
	\item \textbf{Layer insertion:} After $50, 100, \ldots, 400$ epochs.
	\item \textbf{Number of runs:} 1.
	\item \textbf{Decrease of learning rate after layer insertion:} None.
	\item \textbf{Smoothing in Figures:} None.
\end{itemize}

\paragraph{Experiment 12 (CNN)}

\begin{itemize}
	\item \textbf{Figures:} \cref{figure:when-to-insert-layers-cnn}
	\item \textbf{Data set:} CIFAR-10.
	\item \textbf{Data set size} \num{60000}.
	\item \textbf{Train/Test split} \num{50000}/\num{10000}.
	\item \textbf{Data augmentation} RandomCrop, RandomHorizontalFlip, Normalize.
	\item \textbf{Batch size:} 128.
	\item \textbf{Architecture (CNN LI):} Baseline network as described in \cref{figure:CNN-architecture}, then extended by SensLI. Total number of parameters: \num{2674816}, then \num{2711744}.
	\item \textbf{Architecture (CNN1):} Baseline network as described in \cref{figure:CNN-architecture}. Total number of parameters: \num{2674368}
	\item \textbf{Activation function:} \relu.
	\item \textbf{Loss function:} Cross-entropy loss.
	\item \textbf{Optimizer:} SGD with momentum $0.9$, weight decay $0.005$ and learning rate $0.01$.
	\item \textbf{Learning rate schedule:} fixed learning rate.
	\item \textbf{Number of layer insertions by SensLI:} 1.
	\item \textbf{Training epochs total:} 200.
	\item \textbf{Layer insertion:} After $10, 20, 30, \ldots, 80$ epochs.
	\item \textbf{Number of runs:} 1.
	\item \textbf{Decrease of learning rate after layer insertion:} None.
	\item \textbf{Smoothing in Figures:} Moving average with window size 170.
\end{itemize}

\paragraph{Experiment 13 (ResNet)}

\begin{itemize}
	\item \textbf{Figures:} \cref{figure:comparison-with-fixed-architecture-resnets-3lis}, \cref{figure:validity-of-layer-insertion-resnets-3lis}, \cref{figure:vs-random-resnets-3lis}
	\item \textbf{Data set:} Spirals see \cref{figure:illustration-spiral-dataset}
	\item \textbf{Data set size} 600.
	\item \textbf{Train/Test split} 450/150.
	\item \textbf{Data augmentation} None.
	\item \textbf{Batch size:} 45.
	\item \textbf{Architecture (ResNet LI):} ResNet as defined in \cref{eq:ResNet} with 2 hidden layers of width 3, then extended by SensLI 3 times. Total number of parameters: as in \cref{table:parameters-resnet-3lis}.
	\item \textbf{Architecture (ResNet1):} ResNet as defined in \cref{eq:ResNet} with 2 hidden layers of width 3. Total number of parameters: 33.
	\item \textbf{Architecture (ResNet2):} ResNet as defined in \cref{eq:ResNet} with 5 hidden layers of width 3. Total number of parameters: 96.
	\item \textbf{Architecture (ResNet LIother):} as ResNet LI.
	\item \textbf{Architecture (ResNet Random):} as ResNet LI.
	\item \textbf{Activation function:} $\tanh$.
	\item \textbf{Loss function:} Cross-entropy loss.
	\item \textbf{Optimizer:} mini-batch SGD with learning rate $0.1$.
	\item \textbf{Learning rate schedule:} fixed learning rate.
	\item \textbf{Number of layer insertions by SensLI:} 3.
	\item \textbf{Training epochs total:} \num{500}.
	\item \textbf{Layer insertion:} After epoch 100, 200 and 300.
	\item \textbf{Initialization of new layer:} $W_1 \coloneqq 0.8 \, \id_{h_k \times h_k}$ and $b \coloneqq 0$.
	\item \textbf{Number of runs:} 30.
	\item \textbf{Decrease of learning rate after layer insertion:} None.
	\item \textbf{Smoothing in Figures:} None.
\end{itemize}

\begin{table}[h]
	\caption{Number of network parameters during training for the respective networks compared in \cref{figure:validity-of-layer-insertion-resnets-3lis}, \cref{figure:vs-random-resnets-3lis} and \cref{figure:comparison-with-fixed-architecture-resnets-3lis}.}
	\label{table:parameters-resnet-3lis}
	\centering
		\begin{tabular}{rrrrrrr}
			\toprule
			Epochs  & 0--99         & 100--199        & 200--299      & 300--399
			\\
			\midrule
			SensLI  & 33 & 54 & 75 & 96
			\\
			ResNet1     & \multicolumn{4}{c}{33}
			\\
			ResNet2     & \multicolumn{4}{c}{96}
			\\
			\bottomrule
		\end{tabular}
\end{table}

\newpage
\paragraph{Experiment 14 (CNN)}

\begin{itemize}
	\item \textbf{Figures:} \cref{figure:validity-of-layer-insertion-cnn-3lis}, \cref{figure:vs-random-cnn-3lis}
	\item \textbf{Data set:} CIFAR-10.
	\item \textbf{Data set size} \num{60000}.
	\item \textbf{Train/Test split} \num{50000}/\num{10000}.
	\item \textbf{Data augmentation} RandomCrop, RandomHorizontalFlip, Normalize.
	\item \textbf{Batch size:} 128.
	\item \textbf{Architecture (CNN LI):} Baseline network as described in \cref{figure:CNN-architecture}, then extended by SensLI.
	\item \textbf{Architecture (CNN LIother):} Baseline network as described in \cref{figure:CNN-architecture}, then extended by layer with smallest sensitivity.
	\item \textbf{Architecture (CNN Random):} Baseline network as described in \cref{figure:CNN-architecture}, then extended by layer at a random posiiton.
	\item \textbf{Activation function:} \relu.
	\item \textbf{Loss function:} Cross-entropy loss.
	\item \textbf{Optimizer:} Adam with learning rate $0.01$.
	\item \textbf{Learning rate schedule:} ReduceLROnPlateau with patience $10$ and factor $0.5$.
	\item \textbf{Number of layer insertions by SensLI:} 3.
	\item \textbf{Training epochs total:} 200.
	\item \textbf{Layer insertion:} After every 50 epochs.
	\item \textbf{Layer insertion threshold:} as described in \cref{subsection:when-to-insert} with $\tau=1$.
	\item \textbf{Number of runs:} 7.
	\item \textbf{Decrease of learning rate after layer insertion:} None.
	\item \textbf{Smoothing in Figures:} Moving average with window size 170.
\end{itemize}

\printbibliography

\end{document}